\documentclass[11pt]{article}
\usepackage[T1]{fontenc}
\usepackage{graphicx}
\usepackage{amssymb}
\usepackage{amsmath}
\usepackage[ruled]{algorithm}
\usepackage[indLines=true]{algpseudocodex} 

\newcommand{\diag}{{\tt diag}}

\usepackage{mathtools}

\newcommand{\colvec}[2][.8]{%
  \scalebox{#1}{%
    \renewcommand{\arraystretch}{.8}%
    $\begin{bmatrix}#2\end{bmatrix}$%
  }
}

\usepackage{ tikz }
\usetikzlibrary{arrows.meta}
\usetikzlibrary{3d}
\usetikzlibrary{patterns}
\usetikzlibrary{matrix}
\usetikzlibrary{calc}
\usetikzlibrary{positioning}
\usetikzlibrary{decorations.pathreplacing}

\newlength{\lrfigwidth}
\usepackage{stmaryrd}
\usepackage{trimclip}

\makeatletter
\DeclareRobustCommand{\shortto}{\mathrel{\mathpalette\short@to\relax}}
\newcommand{\short@to}[2]{%
  \mkern-3mu
  \clipbox{{.3\width} 0 0 0}{$\m@th#1\vphantom{+}{\shortrightarrow}$}%
  \mkern-3mu
  }
\makeatother

\usepackage{booktabs}
\usepackage{makecell}
\usepackage{multirow}
\usepackage[inline]{enumitem}

\newcommand{\tabletimes}{\mskip3mu\mathsf{x}\mskip3mu}

\newtheorem{theorem}{Theorem}[section]
\newtheorem{proposition}[theorem]{Proposition}

\begin{document}
\title{Computing Linear Regions in Neural Networks with Skip Connections}
%
%
\author{Johnny Joyce \and Jan Verschelde}
\date{University of Illinois at Chicago \\
Department of Mathematics, Statistics, and Computer Science \\
851 S. Morgan St. (m/c 249), Chicago, IL, 6067-7045}
\maketitle              
\begin{abstract}
Neural networks are important tools in machine learning.
Representing piecewise linear activation functions with
tropical arithmetic enables the application of tropical geometry.
Algorithms are presented to compute regions where the neural
networks are linear maps.
Through computational experiments, we provide insights on the
difficulty to train neural networks, in particular on the problems
of overfitting and on the benefits of skip connections.

\end{abstract}

\section{Introduction} \label{sec:tropical-introduction}

A neural network is a composition of neurons,
where each neuron can be represented as a nonlinear function
depending on inputs and parameters, called weights and biases.
The nonlinearity of the network can be understood via
tropical geometry, in particular for networks with ReLU
activation functions, which are piecewise linear.
For such networks, we introduce an algorithm to compute
all linear regions of a neural network.
A linear region of a neural network is a connected region
on which the map defined by the network is linear.
Knowing those linear regions allows for quicker predictions,
as demonstrated by our new caching algorithm.
Our algorithms work for networks with skip connections.
Skip connections add the output of previous layers
to the input of later layers, skipping over the layers in between.

The expository paper~\cite{BHB25} offers promising avenues to
study neural networks. 
The number of linear regions in neural networks was explored
in~\cite{MPCB14} and shown in~\cite{ZNL18} to be exponential 
in the number number of layers, with more results in~\cite{STR18},
though the actual number of linear regions has been observed to be 
much smaller in practice, growing linearly in the number 
of neurons~\cite{HR19}.  The geometry and characteristics
of linear regions was observed when training networks
with different optimization techniques in~\cite{ZW20}.
Tropical geometry was applied to machine learning models in~\cite{MCT21}.
Linear regions have been studied through various perspectives,
such as; by viewing regions as the result of ``foldings'' 
of space \cite{MPCB14},
by viewing regions as affine spline mappings \cite{HBB23}
(which relies on a view of neural networks put forth 
in \cite{BB21}),
and by viewing regions as subsets of tropical maps \cite{ZNL18}. 
The perspective we take in this paper is that of tropical maps.

The mentioned existing literature has studied linear regions
in certain cases, but has not found exact bounds for \emph{all} 
linear regions in any given network.  As such, the problem we solve
in this paper is to compute these exact bounds
of all linear regions in any trained neural network with piecewise
activation functions. Furthermore, we extend those algorithms
to networks with skip connections and obtain insights into
how skip connections produce their beneficial results.

These questions are similar to well-explored problems 
in computational geometry and combinatorial geometry,
where one may examine properties of hyperplane arrangements,
such as the set of all subspaces given by intersections of some 
hyperplanes (this set is called the intersection semilattice)~\cite{Ede87}. 
However, a key difference in our problem is that 
these hyperplanes depend on the input given to the network because
each layer is a function of all layers that precede it.
Certain inputs may cause some combination of neurons to ``fire''
(in the sense that their post-activation function is positive)
on the first layer, while other inputs may lead to other neurons firing,
and the two cases will produce different pre-activation and post-activation
values on all subsequent layers. 
%
Our problem is also similar to \emph{binary space partitioning},
where half-spaces are recursively defined separately for each side 
of some previous hyperplanes, 
resulting in a tree structure~\cite{Tot05}.
However, a key difference is that each layer of a neural network 
introduces multiple hyperplanes at the same time (one for each neuron),
resulting in tree nodes with more than two children.

In this paper, we use tropical geometry to explore linear regions
in neural networks.  In Sect.~\ref{sec:linearregions},
we formulate and present recurrence relations
that can be used to find hyperplanes that bound each linear region
of any given network.  Using these results, we present an algorithm
to find the exact boundaries of every linear region in neural networks.
A walkthrough of our algorithm is presented in Sect.~\ref{sec:examplenetwork}.
We also extend these results and algorithms to networks with arbitrarily
many skip connections in Sect.~\ref{sec:linearregionsskip}.
In Sect.~\ref{sec:visualizelinearregions}, we describe the outcomes
of our computational experiments and visualizations,
observing how patterns of the regions follow patterns in the training data.
Finally, in Sect.~\ref{sec:caching},
we present an algorithm for caching functions produced by 
neural networks on any given linear region, 
allowing for quicker predictions on unseen data.
This algorithm was conceptualized in our previous paper~\cite{JV24}.




All code used in this paper for running algorithms, generating figures,
and performing statistical tests is publicly available
at \cite{tropicalgithub}.
Most of the material for this paper originated in~\cite[Chapter 4]{Joy25}.

\section{Linear Regions of Neural Networks}\label{sec:linearregions}

In this section, we use tropical geometry to introduce an algorithm 
for finding all linear regions of a neural network.

Theorem 5.4 of L.\ Zhang et al.\ \cite{ZNL18} shows that neural networks with piecewise
linear activation functions are equivalent to tropical rational functions. 
They achieve this result by showing that if the inputs to the $\ell$-th
layer of the network are tropical rational functions,
then the output of the layer is also a tropical rational function
(Proposition 5.1 of \cite{ZNL18}, which we also use in 
later sections).  Repeated application of this fact gives us Theorem 5.4
of \cite{ZNL18}, which states the equivalence of neural networks
and tropical rational functions.



Consider a nonlinear neural network
where nonlinearity is introduced only through ReLU activation functions.
This assumption is mild, since more complex piecewise linear functions
can be achieved by adding linear combinations of ReLU functions.
For example, we have $\texttt{LeakyReLU}(x,a)=\texttt{ReLU}(x)-a \cdot \texttt{ReLU}(-x)$, where $a$ specifies the gradient for inputs $x<0$.
With this assumption in mind, 
we have that the output of the first layer of a neural network is given by:
\begin{equation}
   \nu(x)=\texttt{ReLU}(Ax+b)=\max(Ax+b,0),
\end{equation}
where $A$ is the weight matrix of the first layer, 
and $b$ is the corresponding bias. 
\cite{ZNL18} shows that we decompose the matrix $A$ 
as the difference of two matrices, $A := A_{+} - A_{-}$,
where the $(i,j)$-th entry of $A_{+}$ is $\max\{a_{ij},0\}$
and where the $(i,j)$-th entry of $A_{-}$ is $\max\{-a_{ij},0\}$.
This means that the layer $\nu(x)$ can be expressed as the difference 
of two tropical polynomials:
\begin{eqnarray} \label{eq:a_decomposition}
    \nu(x) & = & \max\{A_{+}x+b, A_{-}x\} - A_{-}x \\
           & = & \Big((x^{\otimes A_{+}}\otimes b)
                 \oplus x^{\otimes A_{-}}\Big) \oslash A_{-}x 
\end{eqnarray}
where
\begin{equation}
    x \oplus y  = \max\{x,y\}, \quad x \otimes y = x + y, \quad 
    x \oslash y = x + (-y) = x - y.
\end{equation}

By recursively applying this decomposition of layers, 
\cite{ZNL18} gives the following result, which we use as our starting point:

\begin{proposition}[Proposition 5.1 of L.\ Zhang et al., 2018 \cite{ZNL18}]
\label{prop:zhang51}
Let $A \in \mathbb{Z}^{n_{l+1}\times n_l}$,
$b\in\mathbb{R}^{n_{l+1}}$ be the parameters of the $(\ell+1)$-th layer
of a neural network with piecewise linear activation functions.
Let $\rho^{(\ell)}$ and $\nu^{(\ell)}$ be the pre-activation 
and post-activation output functions of the $\ell$-th layer, respectively.
If the nodes of the $\ell$-th layer are given by tropical rational functions,
\begin{equation}
   \nu^{(\ell)}=F^{(\ell)}(x) \oslash G^{(\ell)}(x)=F^{(\ell)}(x)-G^{(\ell)}(x)
\end{equation}
i.e., each coordinate of $F^{(\ell)}$ and $G^{(\ell)}$
is a tropical polynomial in $x$, then the outputs of the pre-activation
($\rho^{(\ell+1)} \circ \nu^{(\ell)}$) and of the $(\ell+1)$-th layer
($\nu^{(\ell+1)}$) are given by the tropical rational functions
\begin{eqnarray}
   \rho^{(\ell+1)} \circ \nu^{(\ell)}(x)
   & = & H^{(\ell+1)(x)}-G^{(\ell+1)}(x),\\
   \nu^{(\ell+1)}(x)=\sigma \circ \rho^{(\ell+1} \circ \nu^{(\ell)}(x)
   & = & F^{(\ell+1)(x)}-G^{(\ell+1)}(x),
\end{eqnarray}
respectively, where
\begin{eqnarray} \label{eq:zhangrecurrence}
   F^{(\ell+1)}(x) & = & \max\{H^{(\ell+1)}(x), G^{(\ell+1)}(x)\},\\
   G^{(\ell+1)}(x) & = & A_{+}G^{(\ell)}(x) + A_{-}F^{(\ell)}(x),\\
   H^{(\ell+1)}(x) & = & A_{+}F^{(\ell)}(x) + A_{-}G^{(\ell)}(x) + b.
\end{eqnarray}
The tropical forms of these equations using tropical arithmetic
are given in~\cite{ZNL18}.
\end{proposition}
 
From Proposition~\ref{prop:zhang51}, 
we see that $F^{(\ell+1)}(x)=H^{(\ell+1)}(x) \oplus G^{(\ell+1)}(x)$
is the tropical sum of two tropical polynomials. 
Using this, we can determine the linear regions of a neural network 
by examining where $H^{(\ell+1)}$ meets $G^{(\ell+1)}$
at each layer.  However, this is difficult to achieve with the current setup,
because $H^{(\ell+1)}$ and $G^{(\ell+1)}$ depend directly 
on the network's input $x$. 
Therefore, we first need to decompose these recurrence relations 
to minimize their dependence on $x$. 

We can decompose $F$, $G$, and $H$ into the equations
\begin{eqnarray}
    F(x) & = & F_A(x) x + f_b(x) \\
    G(x) & = & G_A(x) x + g_b(x) \\
    H(x) & = & H_A(x) x + h_b(x)
\end{eqnarray}
where $F_A(x),G_A(x),H_A(x) \in \mathbb{R}^{n_{\ell+1}\times n_\ell}[x]$ 
are matrices,
and where $f_b(x)$, $g_b(x)$, $h_b(x) \in \mathbb{R}^{n_{\ell+1}}$ are vectors.

Then the recurrence relations for $F$, $G$, and $H$ given 
by (\ref{eq:zhangrecurrence}) become:
\begin{align}\label{eq:decomposedrecurrence}
   \mathrlap{F_A^{(\ell+1)}(x) = \mathbb{I}_{_{H>G}} H_A^{(\ell+1)}(x) + \mathbb{I}_{_{H\leq G}} G_A^{(\ell+1)}(x),} \nonumber \\
    \mathrlap{f_b^{(\ell+1)}(x) = \mathbb{I}_{_{H>G}} h_b^{(\ell+1)}(x) + \mathbb{I}_{_{H\leq G}} g_b^{(\ell+1)}(x),} \nonumber \\
G_A^{(\ell+1)}(x) & = A_{+}G_A^{(\ell)}(x) + A_{-}F_A^{(\ell)}(x), & & H_A^{(\ell+1)}(x) = A_{+}F_A^{(\ell)}(x) + A_{-}G_A^{(\ell)}(x), \nonumber\\
g_b^{(\ell+1)}(x) & = A_{+}g_b^{(\ell)}(x) + A_{-}f_b^{(\ell)}(x), & & h_b^{(\ell+1)}(x) = A_{+}f_b^{(\ell)}(x) + A_{-}g_b^{(\ell)}(x) + b
\end{align}

where $\mathbb{I}_{_{H>G}}$ evaluates to 1
if $H^{(\ell+1)}(x)>G^{(\ell+1)}(x)$, and 0 otherwise.

This decomposition also requires us to have initial values 
for our recurrences.  To find the initial values, 
consider an identity transformation 
$F^{(0)}(x)=I_{n\times n} \cdot x + \vec{0}$ 
(where $n$ is the input size). 
This gives us that 
$F_A^{(0)}(x) = I_{n\times n}$ 
and 
$f_b^{(0)}(x) = \vec{0}$. 
We can also take $F_A^{(0)}=F_{A+}^{(0)}+F_{A-}^{(0)}$
to be the same type of decomposition into two non-negative matrices
that we used when setting up (\ref{eq:a_decomposition}).
Then (\ref{eq:a_decomposition}) tells us that $F_{A+}^{(0)}=H_A^{(0)}(x)$
and $F_{A-}^{(0)}=G_A^{(0)}(x)$, so our initial values are as follows:
\begin{eqnarray} \label{eq:initialrecurrence}
   F_A^{(0)}(x) & = & I_{n\times n} \qquad
   G_A^{(0)}(x) = 0_{n\times n} \qquad
   H_A^{(0)}(x) = I_{n\times n}, \\
   f_b^{(0)}(x) & = & \vec{0}_{\phantom {n \times n}} \qquad
   g_b^{(\ell+1)}(x) = \vec{0}_{\phantom {\times n}} \qquad
   h_b^{(\ell+1)}(x) = \vec{0}.
\end{eqnarray}
This recurrence relation gives way to the following result:

\begin{theorem} \label{THM:LINEARREGIONS}
Let $f$ be an $L$-layered neural network with ReLU activations 
and with $n_1$, $n_2$, $\dots$, $n_L$ neurons on each layer.
Then any set of vectors $\{ v_\ell \in \{-1,1\}^{n_\ell} \mid \ell\in[L]\}$
uniquely specifies a single linear region of $f$, where the region is given by:
\begin{equation}\label{eq:linearregions}
   \{x \in \mathbb{R}^n \mid \widetilde{H}(x,\ell) \leq \widetilde{G}(x,\ell)
   \mbox{ for all } \ell \in [L] \}
\end{equation}
where
\begin{eqnarray}
   \widetilde{H}(x,\ell) 
   & := & \diag(v_\ell)\left( H_A^{(\ell)}(x) x + h_b^{(\ell)}(x)\right) \\
   \widetilde{G}(x,\ell)
   & := & \diag(v_\ell) \left(G_A^{(\ell)}(x) x + g_b^{(\ell)}(x) \right)
\end{eqnarray}
and where 
$\displaystyle \diag(v_\ell) 
= \sum_{i=1}^{n_\ell} e_i^\intercal v_\ell e_i^\intercal$ 
is the $(n_\ell \times n_\ell)$-matrix where the $(j,j)$-th entry
on the diagonal is given by $(v_i)_j$, 
and where all off-diagonal entries are zero.
    
    Note that some of these regions may be empty. 
    Also note that the set of vectors $v_i \mid_{i \in [L]}$ we construct is often referred to as an \emph{activation pattern}.
\end{theorem}


\noindent {\em Proof.}
We use induction. Let us take any set of vectors $v_i \mid_{i \in [L]}$
and construct the inequalities present in (\ref{eq:linearregions}) 
one layer at a time, starting with the first layer, 
and examining the inequalities we introduce.

\subsubsection*{Base case.}
The first layer introduces inequalities 
$\widetilde{H}(x,1) \leq \widetilde{G}(x,1)$.
Using the initial values given by (\ref{eq:initialrecurrence}),
we obtain:
 
\begin{eqnarray}
   \widetilde{H}(x,1) 
   & = & \diag (v_1) \Big( H_A^{(1)}(x)x + h_b^{(1)}(x) \Big) \\
   & = & \diag (v_1) \Big( \left(A_{+} I_{n_1\times n}
       + A_{-} 0_{n_1\times n} \right)x + \left(A_{+} \vec{0}
       + A_{-} \vec{0} + b\right) \Big) \\
   & = & \diag (v_1) \big( A_{+}x + b \big)
\end{eqnarray}
and
\begin{eqnarray}
   \widetilde{G}(x,1) 
   & = & \diag (v_1) \Big( G_A^{(1)}(x)x + g_b^{(1)}(x) \Big) \\
   & = & \diag (v_1) \Big( \left(A_{+} 0_{n_1\times n}
       + A_{-} I_{n_1\times n}\right)x + \left(A_{+} \vec{0}
       + A_{-} \vec{0} \right) \Big)\\
   & = & \diag (v_1) A_{-}x
\end{eqnarray}
Hence the inequalities introduced are:
\begin{eqnarray}
   \widetilde{H}(x,1) \leq \widetilde{G}(x,1)
   & \iff &
   \diag (v_1) \big( A_{+}x + b \big) \leq \diag (v_1) A_{-}x 
   \\
   & \iff &
   \diag(v_1)\big( Ax + b \big) \leq 0 
   \\
   & \iff &
   (v_1)_i \big( A_i x_i + b_i \big) \leq 0
   \qquad \forall i \in [n_1] \label{eq:appendix_basecaseinequality}
\end{eqnarray}
where $A_i$ is the $i$-th row of $A$.
We have $(v_1)_i \in \{-1,+1\}$ for all $i$, so the inequalities 
in (\ref{eq:appendix_basecaseinequality}) tell us that the choice 
of $-1$ or $+1$ for each element of $v_i$ is equivalent to choosing 
$A_i x_i + b_i \leq 0$ or $A_i x_i + b_i \geq 0$. 
Note that $A_i x_i + b_i$ is the pre-activation output of 
the $i$-th neuron of the first layer. 
By assumption, all activation functions are ReLU, which are linear 
on the region $x\leq 0$ and the region $x \geq 0$ 
(both of these inequalities are non-strict since ReLU is continuous).

Hence, each inequality of (\ref{eq:appendix_basecaseinequality}) 
(i.e.\ each row of the inequality 
$\widetilde{H}(x,1) \leq \widetilde{G}(x,1)$)
limits the region (\ref{eq:linearregions}) exclusively to either 
the positive half or the negative half of each neuron in the first layer.
This completes the base case.

\subsubsection*{Inductive step.}

Take some layer $\ell \in [L]$ of our network and suppose now that 
$\displaystyle \bigcap_{i=1}^{\ell} (\widetilde{H}(x,i)
 \leq \widetilde{G}(x,i))$ 
give the inequalities introduced by our choice 
of $v_i \mid_{i \in [L]}$ for layers $1$ through $\ell$. Then we have:

\begin{eqnarray}
   \widetilde{H}(x,\ell+1) 
   & = &
   \diag (v_{\ell+1}) \Big( H_A^{(\ell+1)}(x)x + h_b^{(\ell+1)}(x) \Big) \\
   & = & 
   \diag (v_{\ell+1}) \Big( \left(A_{+} F_A^{(\ell)}(x) 
   + A_{-} G_A^{(\ell)}(x)\right)x \nonumber \\
   &   & \hphantom{\diag (v_{\ell+1})} + \left(A_{+} f_b^{\ell}(x)
   + A_{-} g_b^{(\ell)}(x) + b \right) \Big)
\end{eqnarray}
and
\begin{eqnarray}
   \widetilde{G}(x,\ell+1) 
   & = &
   \diag (v_{\ell+1}) \Big( G_A^{(\ell+1)}(x)x + g_b^{(\ell+1)}(x) \Big) \\
   & = &
   \diag (v_{\ell+1}) \Big( \left(A_{+} G_A^{(\ell)}(x)
   + A_{-} F_A^{(\ell)}(x)\right)x \nonumber \\
   &  & \hphantom{\diag (v_{\ell+1})} + \left(A_{+} g_b^{\ell}(x)
   + A_{-} f_b^{(\ell)}(x) \right) \Big)
\end{eqnarray}
Subtracting $\widetilde{G}$ from $\widetilde{H}$, we get:
$\widetilde{H}(x,\ell+1) - \widetilde{G}(x,\ell+1)$
\begin{eqnarray}
   & = & \diag (v_{\ell+1}) \Big( \left((A_{+}-A_{-}) F_A^{(\ell)}(x)
       - (A_{+}-A_{-}) G_A^{(\ell)}(x)\right) x \nonumber
   \\
   &   & \qquad \qquad \qquad + \left((A_{+}-A_{-}) f_b^{\ell}(x)
       - (A_{+}-A_{-}) g_b^{(\ell)}(x) + b\right) \Big) 
   \\
   & = & \diag (v_{\ell+1}) \Big( A\left( F^{(\ell)}(x)
       - G^{(\ell)}(x)\right) + b  \Big) 
   \\
   & = & \diag (v_{\ell+1}) \Big( A \nu^{(\ell)}(x)
       + b  \Big) \label{eq:htildeminusgtilde}
\end{eqnarray}
where $\nu^{(\ell)}(x)$
is the post-activation output of the previous layer $\ell$,
as specified in Proposition~\ref{prop:zhang51}. Hence, the inequality $\widetilde{H}(x,\ell+1) \leq \widetilde{G}(x,\ell+1)$
is equivalent to:

\begin{eqnarray}
   & & 
   \hspace{-15mm}
   \widetilde{H}(x,\ell+1) - \widetilde{G}(x,\ell+1) \leq 0 \nonumber \\
   & \iff &
   \diag (v_{\ell+1}) \left( A \nu^{(\ell)}(x) + b \right) \leq 0 
   \\
   & \iff &
   (v_{\ell+1})_i \big( A_i (\nu^{(\ell)} (x))_i + b_i \big) \leq 0,
   \qquad \forall i \in [n_{\ell+1}]. \label{eq:appendix_inductiveinequality}
\end{eqnarray}
 
Using the same reasoning as our base case, 
we see that the choice of each entry of $v_{\ell+1}$
corresponds to choosing whether we take 
$A_i (\nu^{(\ell)} (x))_i + b_i \leq 0$ 
or $ A_i (\nu^{(\ell)} (x))_i + b_i \geq 0$ 
for each neuron. $A_i (\nu^{(\ell)} (x))_i + b_i$ 
is the pre-activation output of the $(\ell+1)$-th layer,
so this choice is equivalent to choosing the positive or negative side 
of the ReLU activation of each neuron in the $(\ell+1)$-th layer.

\subsubsection*{Conclusion of proof.}

Setting 
$\ell=L-1$,
we obtain that the inequalities 
$\displaystyle \bigcap_{i=1}^{L} (\widetilde{H}(x,i) \leq \widetilde{G}(x,i))$
correspond exactly to a single linear region of the network's
final output layer, giving (\ref{eq:linearregions}), 
which completes the proof.

\hfill Q.E.D.


\section{An Algorithm to Find Linear Regions}\label{sec:examplenetwork}

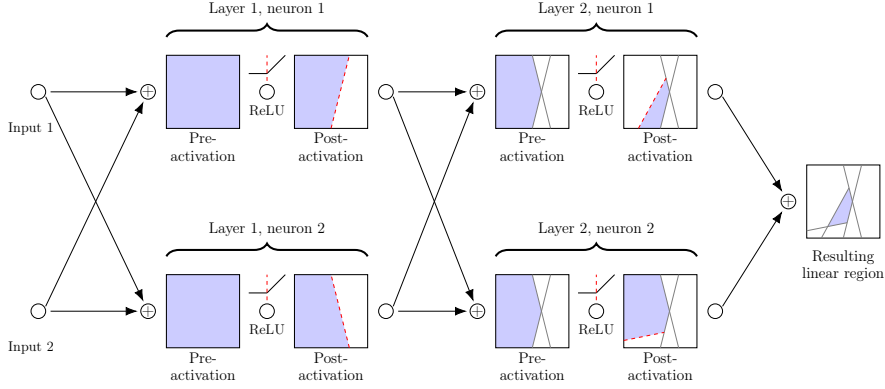
\begin{figure}[htb]
\centering    
\resizebox{0.95\textwidth}{!}{
\newcommand{\ReLUNode}[2]{
    \draw[fill=none] (#1,#2) circle (0.2) node [name=input1,black,xshift=0.2cm] {};
    \node at (#1,#2-0.5) {ReLU};
    \draw[black, thick] (#1-0.5,#2+0.5) -- (#1,#2+0.5);
    \draw[black, thick] (#1,#2+0.5) -- (#1+0.5,#2+1);
    \draw[red, dashed] (#1,#2+0.3) -- (#1,#2+1);
}
\tikzset{
  c/.style={every coordinate/.try},
  box/.style={thick},
  oldhyperplane/.style={gray},
  newhyperplane/.style={thick,red, dashed},
  areashade/.style={
  fill=blue!20, draw opacity=0}
}

\begin{tikzpicture}

    \def\squaresize{2}

    \coordinate (A) at (0,0);
    \coordinate (B) at (\squaresize,0);
    \coordinate (C) at (\squaresize,\squaresize);
    \coordinate (D) at (0,\squaresize);

    \coordinate (neur11a) at (0.5*\squaresize,0);
    \coordinate (neur11b) at (0.75*\squaresize,\squaresize);
    \coordinate (neur12a) at (0.75*\squaresize,0);
    \coordinate (neur12b) at (0.5*\squaresize,\squaresize); 
    
    \coordinate (neur21a) at (0.2*\squaresize, 0);
    \coordinate (neur21b) at (0.75*\squaresize,\squaresize); 
    \coordinate (neur22a) at (0, 0.1*\squaresize);
    \coordinate (neur22b) at (\squaresize,0.3*\squaresize); 

    \coordinate(layer1intersection) at (0.625*\squaresize, 0.5*\squaresize);
    \coordinate(layer2intersection) at (51*\squaresize/178,14*\squaresize/89);
    \coordinate(12intersect21) at (37*\squaresize/64,11*\squaresize/16);
    \coordinate(11intersect22) at (21*\squaresize/38,4*\squaresize/19);
    \coordinate(12intersect22) at (29*\squaresize/42,5*\squaresize/21);

    \draw[fill=none](-4,3+\squaresize/2) circle (0.2) node [name=input1,black,xshift=0.2cm] {}; 
    \draw[fill=none](-4,-3+\squaresize/2) circle (0.2) node [name=input2,black,xshift=0.2cm] {};
    \node at (-4,3+\squaresize/2) [black, xshift=-0.2cm, yshift=-1cm] {Input 1};
    \node at (-4,-3+\squaresize/2) [black, xshift=-0.2cm, yshift=-1cm] {Input 2};

    \draw[fill=none](-1,3+\squaresize/2) circle (0.2) node [name=plus1,black] {\textbf{+}};
    \draw[fill=none](-1,-3+\squaresize/2) circle (0.2) node [name=plus2,black] {\textbf{+}};

    \draw[-{Latex[length=3mm]}](input1.east) -- (plus1);
    \draw[-{Latex[length=3mm]}](input1.south) -- (plus2);
    \draw[-{Latex[length=3mm]}](input2.north) -- (plus1);
    \draw[-{Latex[length=3mm]}](input2.east) -- (plus2);

    \begin{scope}[every coordinate/.style={shift={(-0.5,3)}}]
        \draw[areashade] ([c]A) -- ([c]B) -- ([c]C) -- ([c]D);
        \draw[box] ([c]A) -- ([c]B) -- ([c]C) -- ([c]D) -- cycle;
    \end{scope}
    \begin{scope}[every coordinate/.style={shift={(3,3)}}]
        \draw[areashade] ([c]A) -- ([c]neur11a) -- ([c]neur11b) -- ([c]D);
        \draw[box] ([c]A) -- ([c]B) -- ([c]C) -- ([c]D) -- cycle;
        \draw[newhyperplane] ([c]neur11a) -- ([c]neur11b);
    \end{scope}

    \begin{scope}[every coordinate/.style={shift={(-0.5,-3)}}]
        \draw[areashade] ([c]A) -- ([c]B) -- ([c]C) -- ([c]D);
        \draw[box] ([c]A) -- ([c]B) -- ([c]C) -- ([c]D) -- cycle;
    \end{scope}
    \begin{scope}[every coordinate/.style={shift={(3,-3)}}]
        \draw[areashade] ([c]A) -- ([c]neur12a) -- ([c]neur12b) -- ([c]D);
        \draw[box] ([c]A) -- ([c]B) -- ([c]C) -- ([c]D) -- cycle;
        \draw[newhyperplane] ([c]neur12a) -- ([c]neur12b);
    \end{scope}
    
    \ReLUNode{2.25}{3+\squaresize/2}
    \ReLUNode{2.25}{-3+\squaresize/2}

    \draw[fill=none](5.5,3+\squaresize/2) circle (0.2) node [name=output1,black,xshift=0.2cm] {}; 
    \draw[fill=none](5.5,-3+\squaresize/2) circle (0.2) node [name=output2,black,xshift=0.2cm] {};
    \draw[fill=none](8,3+\squaresize/2) circle (0.2) node [name=plus3,black] {\textbf{+}};
    \draw[fill=none](8,-3+\squaresize/2) circle (0.2) node [name=plus4,black] {\textbf{+}};

    \draw[-{Latex[length=3mm]}](output1.east) -- (plus3);
    \draw[-{Latex[length=3mm]}](output1.south) -- (plus4);
    \draw[-{Latex[length=3mm]}](output2.north) -- (plus3);
    \draw[-{Latex[length=3mm]}](output2.east) -- (plus4);

    \begin{scope}[every coordinate/.style={shift={(8.5,3)}}]
        \draw[areashade] ([c]A) -- ([c]neur11a) -- ([c]layer1intersection) -- ([c]neur12b) -- ([c]D);
        \draw[box] ([c]A) -- ([c]B) -- ([c]C) -- ([c]D) -- cycle;
        \draw[oldhyperplane] ([c]neur11a) -- ([c]neur11b);
        \draw[oldhyperplane] ([c]neur12a) -- ([c]neur12b);
    \end{scope}
    \begin{scope}[every coordinate/.style={shift={(12,3)}}]
        \draw[areashade] ([c]layer1intersection) -- ([c]12intersect21) -- ([c]neur21a) -- ([c]neur21a) --([c]neur11a) -- cycle;
        \draw[box] ([c]A) -- ([c]B) -- ([c]C) -- ([c]D) -- cycle;
        \draw[oldhyperplane] ([c]neur11a) -- ([c]neur11b);
        \draw[oldhyperplane] ([c]neur12a) -- ([c]neur12b);
        \draw[newhyperplane] ([c]neur21a) -- ([c]12intersect21);
    \end{scope}

    \begin{scope}[every coordinate/.style={shift={(8.5,-3)}}]
        \draw[areashade] ([c]A) -- ([c]neur11a) -- ([c]layer1intersection) -- ([c]neur12b) -- ([c]D);
        \draw[box] ([c]A) -- ([c]B) -- ([c]C) -- ([c]D) -- cycle;
        \draw[oldhyperplane] ([c]neur11a) -- ([c]neur11b);
        \draw[oldhyperplane] ([c]neur12a) -- ([c]neur12b);
    \end{scope}
    \begin{scope}[every coordinate/.style={shift={(12,-3)}}]
        \draw[areashade] ([c]D) -- ([c]neur12b) -- ([c]layer1intersection) -- ([c]11intersect22) -- ([c]neur22a) -- cycle;
        \draw[box] ([c]A) -- ([c]B) -- ([c]C) -- ([c]D) -- cycle;
        \draw[oldhyperplane] ([c]neur11a) -- ([c]neur11b);
        \draw[oldhyperplane] ([c]neur12a) -- ([c]neur12b);
        \draw[newhyperplane] ([c]neur22a) -- ([c]11intersect22);
    \end{scope}

    \ReLUNode{11.25}{3+\squaresize/2}
    \ReLUNode{11.25}{-3+\squaresize/2}

    \draw[fill=none](14.5,3+\squaresize/2) circle (0.2) node [name=output3,black,xshift=0.2cm] {}; 
    \draw[fill=none](14.5,-3+\squaresize/2) circle (0.2) node [name=output4,black,xshift=0.2cm] {};
    \draw[fill=none](16.5,0+\squaresize/2) circle (0.2) node [name=plus5,black] {\textbf{+}};
    \draw[-{Latex[length=3mm]}](output3.south) -- (plus5);
    \draw[-{Latex[length=3mm]}](output4.north) -- (plus5);

    \begin{scope}[every coordinate/.style={shift={(17,0)}}]
        \draw[areashade] ([c]layer2intersection) -- ([c]11intersect22) -- ([c]layer1intersection) -- ([c]12intersect21) -- cycle;
        \draw[box] ([c]A) -- ([c]B) -- ([c]C) -- ([c]D) -- cycle;
        \draw[oldhyperplane] ([c]neur11a) -- ([c]neur11b);
        \draw[oldhyperplane] ([c]neur12a) -- ([c]neur12b);
        \draw[oldhyperplane] ([c]neur21a) -- ([c]12intersect21);
        \draw[oldhyperplane] ([c]neur22a) -- ([c]11intersect22);
    \end{scope}

    \draw [decorate, ultra thick,
	decoration = {brace,mirror,amplitude=10pt}] (5,5.5) --  (-0.5,5.5);
    \draw [decorate, ultra thick,
	decoration = {brace,mirror,amplitude=10pt}] (14,5.5) --  (8.5,5.5);
    \node at (2.25,6.25) [black] {\large Layer 1, neuron 1};
    \node at (11.25,6.25) [black] {\large Layer 2, neuron 1};

    \draw [decorate, ultra thick,
	decoration = {brace,mirror,amplitude=10pt}] (5,-0.5) --  (-0.5,-0.5);
    \draw [decorate, ultra thick,
	decoration = {brace,mirror,amplitude=10pt}] (14,-0.5) --  (8.5,-0.5);
    \node at (2.25,0.25) [black] {\large Layer 1, neuron 2};
    \node at (11.25,0.25) [black] {\large Layer 2, neuron 2};

    \node at (0.5,-3.5) [black,align=center] {\large Pre-\\  \large activation};
    \node at (4,-3.5) [black,align=center] {\large Post-\\  \large activation};
    \node at (9.5,-3.5) [black,align=center] {\large Pre-\\  \large activation};
    \node at (13,-3.5) [black,align=center] {\large Post-\\  \large activation};

    \node at (0.5,2.5) [black,align=center] {\large Pre-\\  \large activation};
    \node at (4,2.5) [black,align=center] {\large Post-\\  \large activation};
    \node at (9.5,2.5) [black,align=center] {\large Pre-\\  \large activation};
    \node at (13,2.5) [black,align=center] {\large Post-\\  \large activation};
    
    \node at (18,-0.75) [black,align=center] {\large Resulting\\ \large linear region};
    
\end{tikzpicture}
}
\caption[A visual representation of Algorithm~\ref{algorithm:findlinearregions}
         on a simple neural network.]
{A visualization of the walkthrough of 
Algorithm~\ref{algorithm:findlinearregions}
given in Sect.~\ref{sec:examplenetwork}. 
For each neuron with ReLU activation, 
we split the input space in two by introducing a hyperplane 
and by selecting a half space with $(v_\ell)_i$. 
The newly-introduced hyperplanes at each neuron (as given by Theorem~\ref{THM:LINEARREGIONS}) are highlighted 
in red, and the shaded blue area represents the selected 
intersection of half spaces that comprise the linear region.}
\label{fig:linearregionsnetwork}
\end{figure}

Theorem~\ref{THM:LINEARREGIONS} now facilitates 
Algorithm~\ref{algorithm:findlinearregions} for finding the 
linear region specified by any set of vectors $v_\ell \mid_{l\in[L]}$, 
without the need to provide a direct input $x$ to the neural network.
A visual representation of Algorithm~\ref{algorithm:findlinearregions} 
is given in Fig.~\ref{fig:linearregionsnetwork}.

\begin{algorithm}[hbt] \caption{Method to find linear regions of neural networks, as in Theorem~\ref{THM:LINEARREGIONS}.}\label{algorithm:findlinearregions}
\begin{minipage}{0.98\textwidth}
\hfuzz=5pt 
\begin{algorithmic}[1]
\Function{FindLinearRegion}{$v_1$, $\dots$, $v_L$}
    \State $F_A^{(0)}, G_A^{(0)}, H_A^{(0)}, f_b^{(0)}, g_b^{(0)}, h_b^{(0)} \gets$
    Initial values laid out in (\ref{eq:initialrecurrence}) \label{algline:initialvalues}
    \State $S \gets $ Empty array  \Comment{To be filled with inequalities that define the linear region.}
    \For{$l=0,\dots,L-1$}
        \State $A \gets $ weight matrix of layer $\ell+1$
        \State $b \; \gets $ bias vector of layer $\ell+1$
        \State $G_A^{(l+1)}, H_A^{(l+1)}, g_b^{(\ell+1)}, h_b^{(\ell+1)} \gets
	$ Values for the next layer laid out in (\ref{eq:decomposedrecurrence}) 
    \label{algline:recurrence}
        \State $F_A^{(\ell+1)} \gets $ a blank matrix to be filled out \label{algline:fstart}
        \State $f_b^{(\ell+1)} \, \gets $ a blank vector to be filled out 
        \For{$i=1,\dots,l$}
            \State $\vec{\alpha} \gets$ $i$-th row of $H_A^{(\ell+1)}-G_A^{(\ell+1)}$
            \State $\beta \gets$ $i$-th entry of $h_b^{(\ell+1)}-g_b^{(\ell+1)}$
            \If{$(v_{l+1})_i == +1$}
                \State Append the inequality $\vec{\alpha}\cdot x + \beta \leq 0$ to $S$ \label{algline:ineq1}
                \State $i$-th row of $F_A^{(\ell+1)}$ \; $\gets$ $i$-th row of $G_A^{(\ell+1)}$
                \State $i$-th entry of $f_b^{(\ell+1)}$ $\gets$ $i$-th entry of $g_b^{(\ell+1)}$
            \Else
                \State Append the inequality $-\vec{\alpha}\cdot x - \beta \leq 0$ to $S$ \label{algline:ineq2}
                \State $i$-th row of $F_A^{(\ell+1)}$ \; $\gets$ $i$-th row of $H_A^{(\ell+1)}$
                \State $i$-th entry of $f_b^{(\ell+1)}$ $\gets$ $i$-th entry of $h_b^{(\ell+1)}$
            \EndIf
        \EndFor \label{algline:fend}
        \If{the region given by the intersection of elements of $S$ is empty}
            \State \Return Message indicating that the region is empty,

	    and specify the current index $\ell$ where we reached an empty set
        \EndIf
    \EndFor
    \State \Return S
\EndFunction
\end{algorithmic}
\end{minipage}
\end{algorithm}

The entries of vectors $v_\ell \mid_{\ell\in[L]}$ uniquely correspond
to a single linear region, which may be empty or nonempty.
This means that Algorithm~\ref{algorithm:findlinearregions} allows us 
to find \emph{all} linear regions. 
A brute-force method for doing this would be to enumerate all possible
assignments of $-1$ and $+1$ to entries in~$v_\ell$, which would
have exponential complexity in the total number
of neurons, since the number of assignments 
is $\displaystyle 2^{\sum_{\ell=1}^{L} n_\ell}$. Instead, we can reduce the complexity by using a tree traversal approach, which is laid out in Algorithm~\ref{algorithm:dfs}.

\begin{algorithm}[hbt] \caption{A tree traversal approach to finding all 
linear regions. 
Initial call is \textsc{Traverse}(0,$\varnothing$).}\label{algorithm:dfs}
\begin{algorithmic}[1]
\hfuzz=5pt 
\State results $ \gets $ a global array, accessible within nested calls
\Function{Traverse}{depth, $v$} 

\Comment{$v$ is some set of elements $\{v^{(1)},\dots,v^{(k)}\}$. 
Initially, $v$ is empty.}
    \If{depth $== L$}
        \State msg $\gets$ \textsc{FindLinearRegion}$(v)$
	\Comment{Call algorithm~\ref{algorithm:findlinearregions}.}
        \If{msg shows an empty region}
            \State \Return index of failure
        \Else
            \State Append msg containing linear region to global array of results
            \State \Return nothing
        \EndIf
    \Else
        \For{each possible assignment of $-1$ and $+1$ to entries of $v_{\mbox{depth}}$}
            \State $v' \gets$ current assignment of $-1$ and $+1$
            \State msg $\gets$ \textsc{Traverse}(depth$+1$, $v \cup v'$) 
            \Comment{Recursive call.}
            \If{msg contains index of a failure, and current depth $>$ index of failure}
                \State \Return index of failure
            \EndIf
        \EndFor
    \EndIf
\EndFunction
\end{algorithmic}
\end{algorithm}

Let us now follow a concrete walkthrough of 
Algorithm~\ref{algorithm:findlinearregions} on a small example network. 
For this example, consider a network with a 2-dimensional 
input layer, 2 hidden layers with 2 neurons per layer, 
and a 1-dimensional output layer. 
Suppose that the weights and biases of each hidden layer 
in the network are:

\begin{equation}
   A^{(1)} 
   = \left[
        \begin{array}{rr}
           -4 & ~1\\
           -4 & ~-1
        \end{array}
     \right], \quad
   b^{(1)} 
   = \left[
       \begin{array}{r}
          2 \\ 3
       \end{array}
     \right], \quad
   A^{(2)}
   = \left[
        \begin{array}{rr}
              -8 & ~3 \\
           -\frac{21}{4} & ~\frac{19}{4}
        \end{array}
     \right], \quad
   b^{(2)}
   = \left[
       \begin{array}{r}
         -4 \\ 1
       \end{array}
     \right].
\end{equation}
Let us assume that the 1-dimensional output layer
does not have an activation function, so we can ignore it and instead 
focus on the two hidden layers. 

Take the arbitrary truth assignment 
$v_1=\colvec{1\\1}$,
$v_2=\colvec{1\\-1}$.
Each entry corresponds to whether choosing either the positive 
or negative half of the activation function for each neuron 
in the network, so we want to find the linear region 
that corresponds to this truth assignment.

Using Algorithm~\ref{algorithm:findlinearregions}, 
on line \ref{algline:initialvalues}, we first set the initial
values $F_A^{(0)}$, $G_A^{(0)}$, $H_A^{(0)}$, $f_b^{(0)}$,
$g_b^{(0)}$, $h_b^{(0)}$ to the initial values given 
by (\ref{eq:initialrecurrence}).
Then, on line \ref{algline:recurrence} of the algorithm,
we apply the recurrence relations given by (\ref{eq:decomposedrecurrence}),
giving the following:
 
\begin{equation}
   G_A^{(1)}
   = \left[
        \begin{array}{rr}
           4 & ~0\\
           4 & ~1
        \end{array}
     \right], \quad
   g_b^{(1)}
   = \left[
        \begin{array}{r}
           0 \\ 0
        \end{array}
     \right], \quad
   H_A^{(1)} 
   = \left[
       \begin{array}{rr}
           0 & ~1 \\
           0 & ~0
       \end{array}
     \right], \quad
   h_b^{(1)}
   = \left[
       \begin{array}{r}
          2 \\ 3
       \end{array}
     \right].
\end{equation}
 

Next, lines \ref{algline:fstart}-\ref{algline:fend} 
of Algorithm~\ref{algorithm:findlinearregions} allow us 
to construct $F_A^{(1)}$ and $f_b^{(1)}$ by using $v_1$ 
to select from $G_A^{(1)}$ and $g_b^{(1)}$,
or from $H_A^{(1)}$ and $h_b^{(1)}$ for each row.
Since we have 
$v_1=\colvec{1\\1}$,
we select $G_A^{(1)}$ and $g_b^{(1)}$ for both rows, giving:
 
\begin{equation}
   F_A^{(1)}
   = \left[
       \begin{array}{rr}
          4 & ~0\\
          4 & ~1
       \end{array}
     \right], \quad
   f_b^{(1)}
   = \left[
       \begin{array}{r}
          0 \\ 0
       \end{array}
     \right].
\end{equation}

The entries of $v_1$ correspond to selecting 
a half space for each neuron on the first layer,
which produce a linear region when all such half spaces 
are intersected. 
These half spaces are specified by the inequalities 
on lines \ref{algline:ineq1} and \ref{algline:ineq2} 
of Algorithm~\ref{algorithm:findlinearregions}. 
Hence, at the end of the first layer, 
the linear region is given by:

\begin{equation} \label{eq:examplelinearbound1}
    \Big(
      \left[
        \begin{array}{rr}
          4 & ~0 \\
          4 & ~1
        \end{array}
      \right]
      - 
      \left[
        \begin{array}{rr}
          0 & ~1 \\
          0 & ~0
        \end{array}
      \right]
    \Big)
    \left[
      \begin{array}{r}
        x \\ y
      \end{array}
    \right]
    +
    \Big(
     \left[
      \begin{array}{r}
         0 \\ 0
      \end{array}
     \right]
     -
     \left[
      \begin{array}{r}
         2 \\ 3
      \end{array}
     \right]
    \Big)
    \leq
    \left[
      \begin{array}{r}
         0 \\ 0
      \end{array}
    \right].
\end{equation}

Next, we must apply the same recurrence relations 
to the second layer, giving:

\begin{equation}
   G_A^{(2)} 
   = \left[
       \begin{array}{rr}
          12 & ~11 \\
          19 & ~10
       \end{array}
     \right], \quad
   g_b^{(2)}
   = \left[
       \begin{array}{r}
          16 \\
          \frac{21}{2}
       \end{array} 
     \right], \quad
   H_A^{(2)}
   = \left[
       \begin{array}{rr}
          32 & ~0 \\
          21 & ~0
       \end{array}
     \right], \quad
   h_b^{(2)}
   = \left[
       \begin{array}{r}
          12 \\ \frac{23}{2}
       \end{array}
     \right].
\end{equation}
For layer 2, our truth assignment is 
$\colvec{1\\-1}$,
so we select $F_A^{(2)}$ and $f_b^{(2)}$ 
by taking a mixture of both $G_A^{(2)}$ 
\& $g^{(2)}$, and $H_A^{(2)}$ \& $h_b^{(2)}$. This gives us:

\begin{equation}
   F_A^{(2)}
   = \left[
       \begin{array}{rr}
          32 & ~0 \\
          19 & ~10
       \end{array}
     \right], \quad
   f_b^{(2)}
   = \left[
       \begin{array}{r}
          12 \\ \frac{21}{2}
       \end{array}
     \right].
\end{equation}
Finally, we obtain the bounds that this layer
introduces to our linear region by again taking
the inequalities on lines \ref{algline:ineq1} 
and \ref{algline:ineq2} of Algorithm~\ref{algorithm:findlinearregions}, giving:
 
\begin{equation} \label{eq:examplelinearbound2}
   \Big(
     \left[
       \begin{array}{rr}
          12 & ~11 \\
          21 & ~0
       \end{array}
     \right]
     -
     \left[
       \begin{array}{rr}
          32 & ~0 \\
          19 & ~10
       \end{array}
     \right]
   \Big)
   \left[
     \begin{array}{r}
        x \\ y
     \end{array}
   \right]
   +
   \Big(
     \left[
       \begin{array}{r}
          16 \\ \frac{23}{2}
       \end{array}
     \right]
     - 
     \left[
       \begin{array}{r}
          12 \\ \frac{21}{2}
       \end{array}
     \right]
   \Big) 
   \leq
   \left[
     \begin{array}{r}
        0 \\ 0
     \end{array}
   \right].
\end{equation}
Note that the top rows of all matrix subtractions
correspond to taking $G_A^{(2)}-H^{(2)}_A$,
while the bottom rows correspond to $H_A^{(2)}-G_A^{(2)}$. 
The inequalities given by (\ref{eq:examplelinearbound2}) 
can be expressed as $\{(x,y)\in \mathbb{R}^2 \mid -20x+11y\leq -4, 2x-10y \leq -1\}$.

Taking the intersection of the half spaces given 
by (\ref{eq:examplelinearbound1}) and (\ref{eq:examplelinearbound2}), 
we obtain our final linear region, which is given by:

\begin{equation} \label{eq:examplelinearregion}
\left\{ \; (x,y)\in \mathbb{R}^2 \;\;\middle\vert\;\;
{
   \begin{array}{rclccrcl}
      4x-y     & \leq & 2,  & &   & 4x+y  & \leq & 3 \\
      -20x+11y & \leq & -4, & &   & x-10y & \leq & -1
   \end{array} 
}\;
\right\}.
\end{equation}

Fig.~\ref{fig:linearregionsnetwork} shows a visualization of this example,
where the constraints imposed on the linear region by each neuron are shaded
in blue for the region $[0,1]\times[0,1]\subset \mathbb{R}^2$.
The resulting linear region shown on the right side of the figure exactly 
matches the constraints given by (\ref{eq:examplelinearregion}).

If we were to repeat this process for all possible truth assignments
$v_1$ and $v_2$, we would obtain all linear regions of the network.
Algorithm~\ref{algorithm:dfs} allows us to efficiently explore all
such truth assignments. 

\section{Linear Regions in Networks With Skip Connections}\label{sec:linearregionsskip}

Using the results of Sect.~\ref{sec:linearregions}, we can also find linear regions of networks with skip connections 
--- we need only update our recurrence relations.
In Sect.~\ref{sec:linearregions}, our recurrences relied on the fact that 
the post-activation output $\nu$ of the $(\ell+1)$-th layer was given by 
$\nu^{(\ell+1)}(x)=\max\{A\nu^{(\ell)}(x)+b,0\}$. 
To account for skip connections, suppose that for any layer $\ell+1$, there is some subset $K\subset [\ell]$
of layers preceding layer $\ell+1$ that have a skip connection leading 
to layer $\ell+1$.
Then the input to layer $\ell+1$ 
is $\displaystyle \nu^{(\ell)}(x)+\sum_{k \in K} \nu^{(k)}(x)$.
Therefore, the recurrence relation for the post-activation output 
of layer $\ell+1$ is:

\begin{eqnarray}
   \nu^{(\ell+1)}(x)
   & = &
   \max\Big\{A\Big(\nu^{(\ell)}(x)+\sum_{k\in K}\nu^{(k)}(x)\Big)+b,0\Big\} 
   \\
   & = & \max\Big\{A_{+}\nu^{(\ell)}(x)+b+A_{+} \sum_{k\in K}\nu^{(k)}(x), A_{-}\nu^{(l)}(x) \nonumber
   \\
   &   & \qquad \quad +A_{-}\sum_{k\in K}\nu^{(k)}(x)\Big\} 
   - \Big(A_{-}\nu^{(l)}(x)+A_{-}\sum_{k\in K}\nu^{(k)}(x)\Big) \label{eq:recurrenceskip}
\end{eqnarray}
We can express $\nu^{(k)}(x)$ as $(F^{(k)}(x)-G^{(k)}(x))$,
giving:

\begin{eqnarray}\label{eq:recurrenceskip2}
   F^{(\ell+1)}(x) & = & \max\{H^{(\ell+1)}(x), G^{(\ell+1)}(x)\} \\
   G^{(\ell+1)}(x) & = & A_{+}G^{(\ell)}(x) + A_{-}F^{(\ell)}(x)
   + \underline{A_{-}\sum_{k\in K}\Big(F^{(k)}(x)-G^{(k)}(x)\Big)} \\
   H^{(\ell+1)}(x) & = & A_{+}F^{(\ell)}(x) + A_{-}G^{(\ell)}(x) + \underline{A_{+}\sum_{k\in K}\Big(F^{(k)}(x)-G^{(k)}(x)\Big)} + b \; \; \;
\end{eqnarray}
Here, the altered parts are underlined for emphasis. 
Lastly, we need recurrence relations for $H_A, h_b, G_A$, and $g_b$, 
since this allows us to calculate the boundaries of linear regions.
These relations are as follows:
\begin{eqnarray} \label{eq:decomposedrecurrenceskip}
   G_A^{(\ell+1)}(x)
   & = & A_{+}G_A^{(\ell)}(x) + A_{-}F_A^{(\ell)}(x)
         + \underline{A_{-}\sum_{k\in K}\Big(F_A^{(k)}(x)-G_A^{(k)}(x)\Big)} \\
   g_b^{(\ell+1)}(x)
   & = & A_{+}g_b^{(\ell)}(x) + A_{-}f_b^{(\ell)}(x)
         + \underline{A_{-}\sum_{k\in K}\Big(f_b^{(k)}(x)-g_b^{(k)}(x)\Big)} \\
   H_A^{(\ell+1)}(x)
   & = & A_{+}F_A^{(\ell)}(x) + A_{-}G_A^{(\ell)}(x) 
         + \underline{A_{+}\sum_{k\in K}\Big(F_A^{(k)}(x)-G_A^{(k)}(x)\Big)} \\
   h_b^{(\ell+1)}(x)
   & = & A_{+}f_b^{(\ell)}(x) + A_{-}g_b^{(\ell)}(x)  + \underline{A_{+}\sum_{k\in K}\Big(f_b^{(k)}(x)-g_b^{(k)}(x)\Big)} + b \; \;
\end{eqnarray}
and $F_A$ and $f_b$ are unchanged from before.
The conclusion of this section is that the relations given 
in (\ref{eq:decomposedrecurrenceskip}) can be used 
in Algorithms~\ref{algorithm:findlinearregions}~and~\ref{algorithm:dfs} 
to find all linear regions of neural networks with arbitrarily many skip connections.

\section{Visualization of Linear Regions in Neural Networks}\label{sec:visualizelinearregions}

In this section, we use 
Algorithms~\ref{algorithm:findlinearregions}~and~\ref{algorithm:dfs}
to plot \emph{all} linear regions in some small neural networks,
and we analyze their appearance.

\begin{figure}[hbtp]
\centering
\setlength{\lrfigwidth}{0.3333\textwidth}
\setlength\tabcolsep{0pt} 
\renewcommand{\arraystretch}{2.5} 
\begin{tabular}{ccc}
\includegraphics[width=\lrfigwidth]{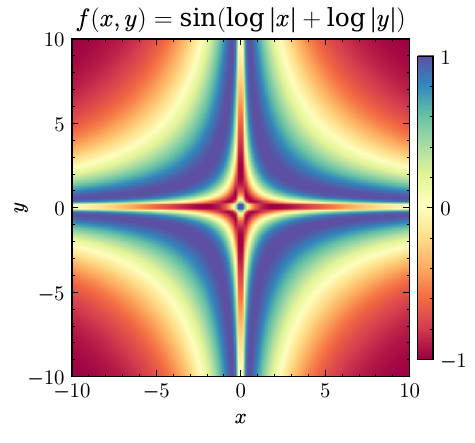} & 
\includegraphics[width=\lrfigwidth]{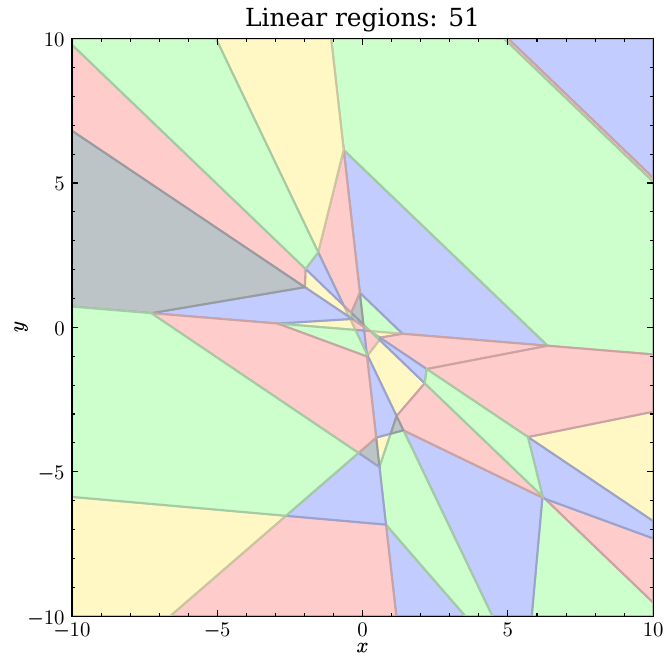} &
\includegraphics[width=\lrfigwidth]{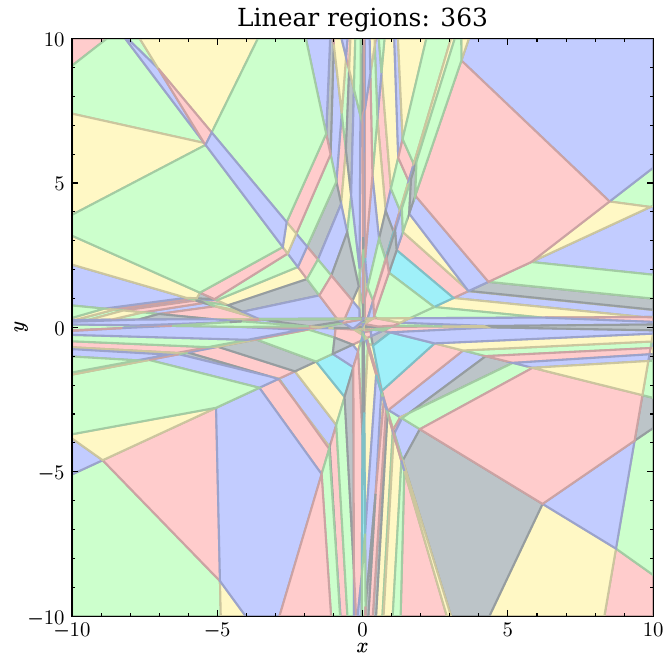} \\
\includegraphics[width=\lrfigwidth]{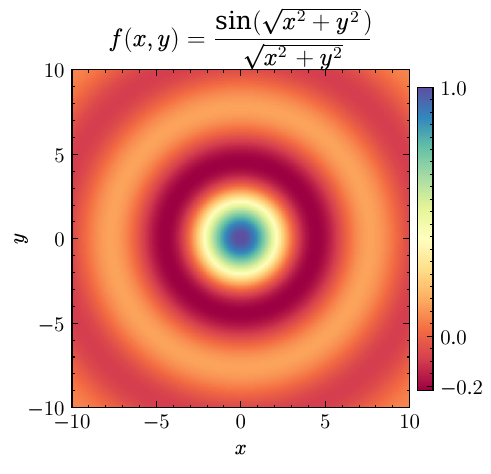} & 
\includegraphics[width=\lrfigwidth]{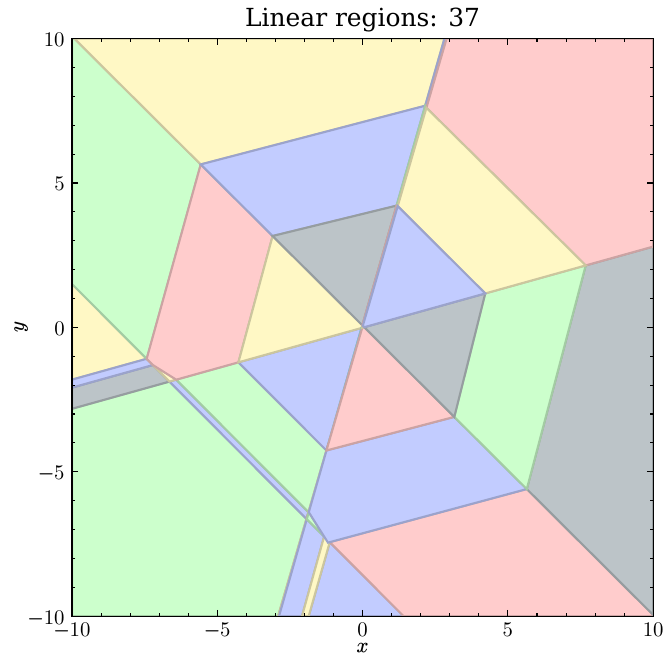} &
\includegraphics[width=\lrfigwidth]{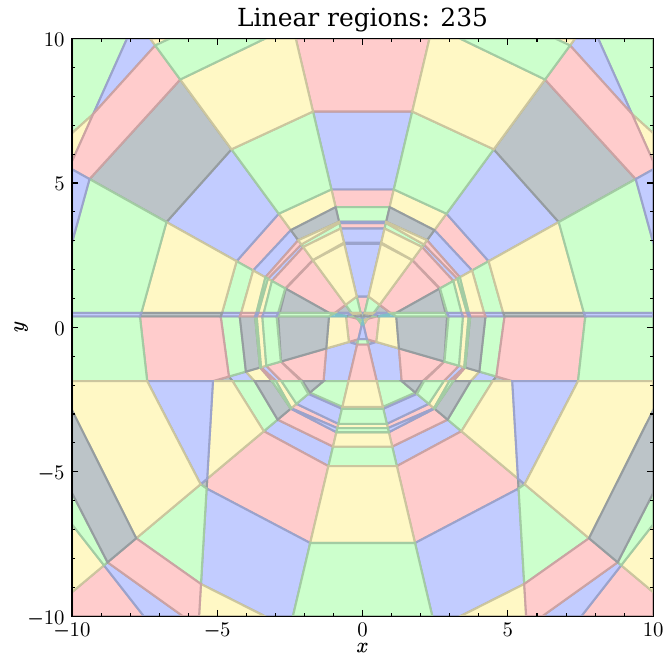} \\
\includegraphics[width=\lrfigwidth]{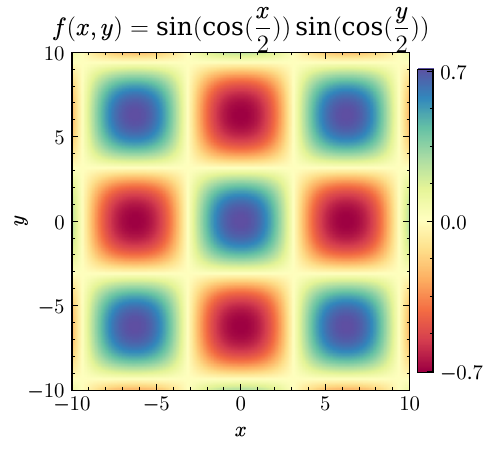} & 
\includegraphics[width=\lrfigwidth]{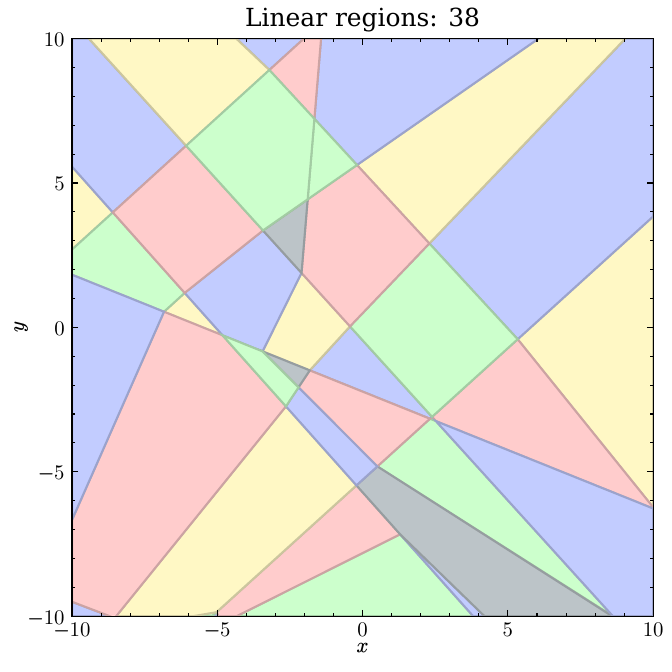} &
\includegraphics[width=\lrfigwidth]{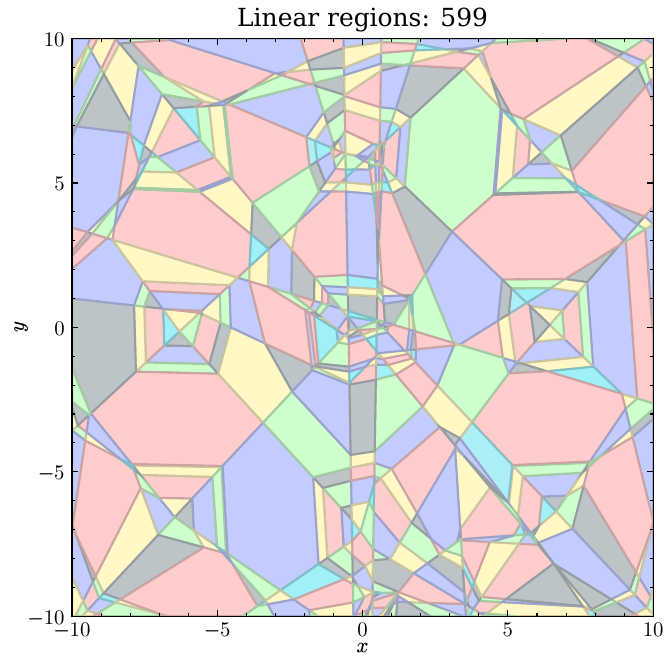} \\
\includegraphics[width=\lrfigwidth]{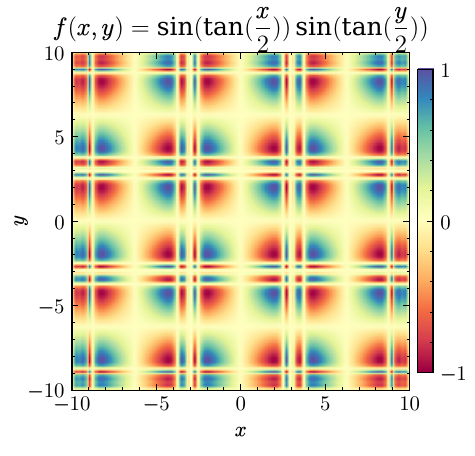} & 
\includegraphics[width=\lrfigwidth]{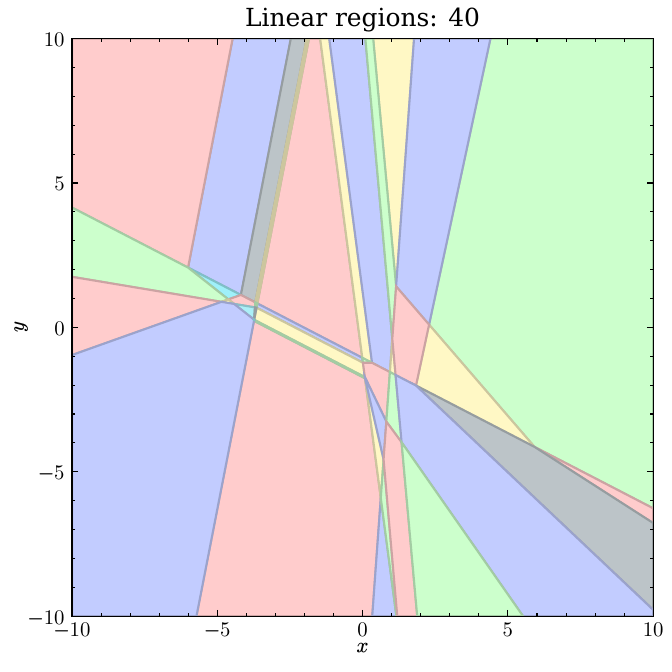} &
\includegraphics[width=\lrfigwidth]{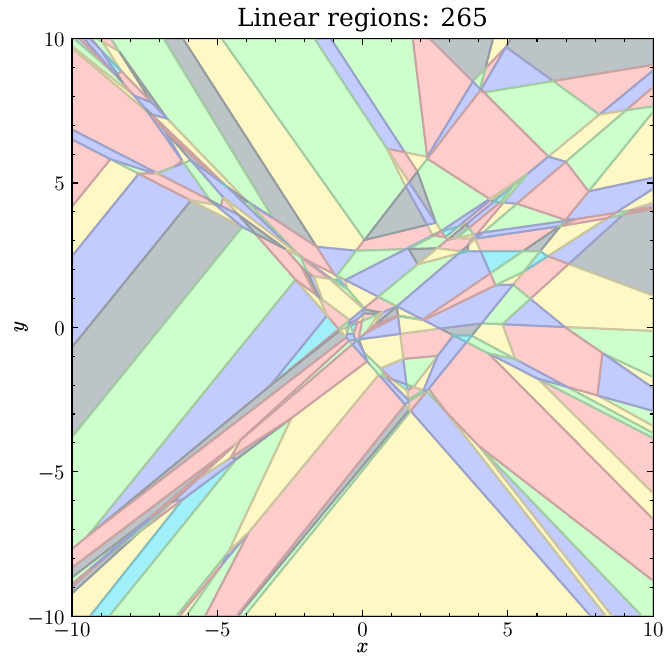} \\
\end{tabular}
\caption[Plots of linear regions of neural networks trained on various functions.]
{Linear regions of neural networks after 20,000 training epochs. 
Each row $i$ corresponds to a function $\phi_i$ in (\ref{eq:plotfunctions}). 
Colors are arbitrary.
\textit{Left column:} Plot of function given as training data. 
\textit{Middle column:} Linear regions of 2-layer neural network.
\textit{Right column:} Linear regions of 5-layer neural network. }
\label{fig:linearregionvisualization}
\end{figure}

For easy visualization, we consider regression problems on 
neural networks with 2 scalar input variables $x$ and $y$,
and 1 scalar output variable $z$.
We perform experiments with each of the following functions:

\begin{eqnarray} \label{eq:plotfunctions}
   \phi_1(x,y) & = & \sin(\log\left|x\right|+\log\left|y\right|) \\
   \phi_2(x,y) & = & \sin(\sqrt{x^2+y^2}) \div \sqrt{x^2+y^2} \\
   \phi_3(x,y) & = & \sin\big(\cos(x/2)\big)
                     \sin\big(\cos(y/2)\big) \\
   \phi_4(x,y) & = & \sin\big(\tan(x/2)\big)
                     \sin\big(\tan(y/2)\big) \label{eq:plotfunctionslast}
\end{eqnarray}
$\phi_1$, $\phi_2$, and $\phi_3$ are examples of functions 
that a neural network can be expected to model well.
On the other hand, $\phi_4$ is given as an unreasonably difficult 
function for a small network to model, which we expect to lead 
to overfitting, allowing us to observe the linear regions in such a scenario.
These functions are plotted in the left column 
of Fig.~\ref{fig:linearregionvisualization}.
For each of these $\phi$ functions, we train simple neural networks with 
the Adam optimizer 
and with learning rate $\eta=0.01$ for 20,000 epochs.
Each training instance $(x,y)$ is sampled from $[-10,10]\times[-10,10]$ 
on a uniform grid at intervals of 0.1. 
For each function, two networks are trained. 
Both have ReLU activation on all hidden nodes, and both have 5 neurons 
per layer
The only difference between the two networks is that one has 2 hidden layers, 
while the other has 5 hidden layers, allowing us to observe the effects of adding more layers.
This is not the first experiment to produce visualizations of linear regions, 
as there is precedence from Mont\'{u}far et al.\ \cite{MPCB14},
Hanin and Rolnick \cite{HR19}, and X. Zhang and Wu \cite{ZW20}, 
who have all made similar visualizations. However, we believe that this is 
the first experiment that captures \emph{all} distinct linear regions 
of a neural network and provides exact boundaries for these regions.

The resulting linear regions of the trained networks are shown
in Fig.~\ref{fig:linearregionvisualization}. From this,
we make the following observations:

\begin{itemize}[wide]
\item For $\phi_1$ and $\phi_2$, the 5-layer network partitioned
the input region into regions that mimic the patterns in the training data. 
The 2-layer network also produced visual patterns 
matching the training data in less detail,
suggesting that deeper networks may be able to position boundaries
of linear regions more effectively than shallow networks,
which partially explains their increased effectiveness
(among many other known explanations).

\item $\phi_3$ had contours that were more difficult to model 
than $\phi_1$ and $\phi_2$, but the patterns produced 
by the 5-layer network are still sensible. 
The deeper nature of the 5-layer network over the 2-layer network was necessary to
capture this pattern, where early layers produced long, sweeping hyperplanes surrounding each peak and valley,
and successive layers captured the smaller radial patterns.
Our network only had 5 neurons per layer, yet 9 such patterns were made,
suggesting that the divisions produced in early layers allow the input space to be ``folded'' in such a way that patterns produced by later layers are repeated
across the space. This folding analogy is detailed by Mont\'{u}far et al.\ \cite{MPCB14}.
    
\item On the other hand, for $\phi_4$ (which
we expect to induce overfitting), the networks produced clusters of tiny 
linear regions near the center, with highly non-uniform regions near the edges. 
This shows us that a neural network that 
overfits on a task may start dividing the input space randomly via
early layers, before subdividing resulting small regions further.
This is consistent with an intuitive understanding of overfitting,
which is that it occurs when too much emphasis is placed on noise in training data. 
\end{itemize}


\section{Comparison With and Without Skip Connections}\label{sec:linearregions-skipvsnoskip}

In this section, we compare the number of linear regions 
in neural networks with and without skip connections. 
We start in Subsect.~\ref{subsec:skipvsnoskip-training} 
by making this comparison throughout training,
where we find that neural networks with skip connections have more linear regions than regular networks, 
even with no training. 
Given this unexpected result, we verify it 
with hypothesis tests in Subsect.~\ref{subsec:statisticaltest-randominit}.

\subsection{Comparison Throughout Training}\label{subsec:skipvsnoskip-training}

We train two types of neural networks on each of the functions 
$\phi_1$, $\phi_2$, $\phi_3$, $\phi_4$ given in (\ref{eq:plotfunctions}-\ref{eq:plotfunctionslast}). 
We create the training data in the same way as 
in Sect.~\ref{sec:visualizelinearregions}, though we also add a testing set,
where we sample at the same interval as the training set, and offset 
the samples by 0.1 (e.g.\ the first few $x$ coordinates in the training 
set are -$10$, $-9.8$, $-9.6$, $\dots$, while the first few $x$ coordinates 
of the testing set are $-9.9$, $-9.7$, $-9.5$, $\dots$).
Both types of networks have 5 hidden layers, with 4 neurons per layer.
One type of network does not have skip connections,
while the other has a skip connection from the first layer
to the third layer, and from the second layer to the fourth layer.
For each type of network, we perform 10 trials on each function
$\phi_1$ through $\phi_4$, and take the overall mean across all trials
and functions of the number of linear regions throughout training.
We also record the loss to validate our methodology.

\begin{figure}[hbt]
\centering
\renewcommand{\arraystretch}{0.25} 
\begin{tabular}{c}
\includegraphics[width=0.75\textwidth]{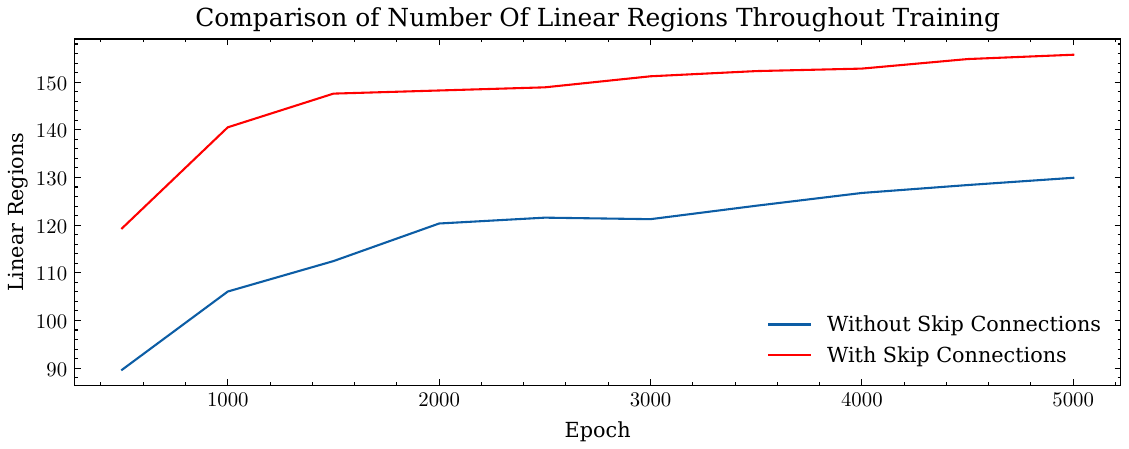} \\
\includegraphics[width=0.75\textwidth]{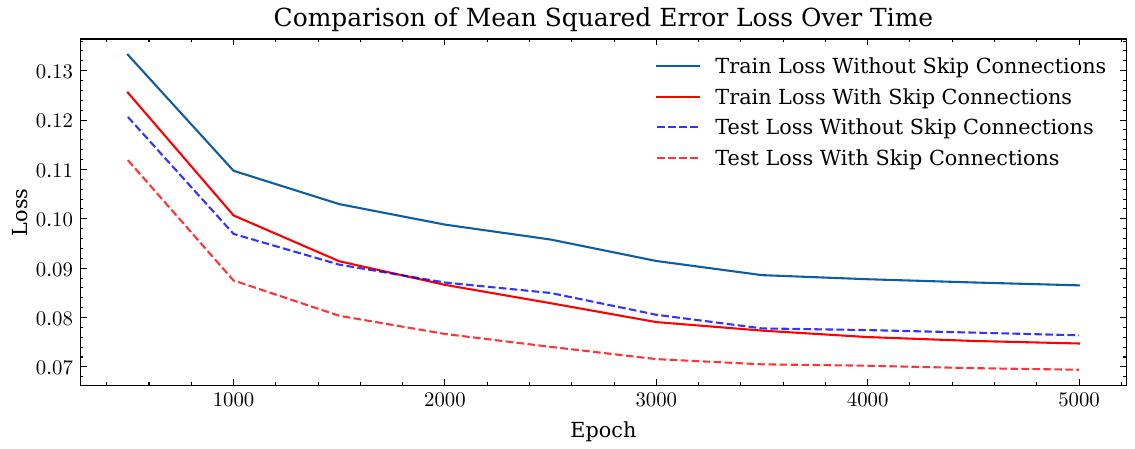}
\end{tabular}
\caption[Comparison of neural networks with and without skip connections.]
{Comparison of neural networks with and without skip connections throughout
training.  Networks were trained on functions $\phi_1$, $\phi_2$, $\phi_3$,
$\phi_4$, and the mean was taken over 10 trials of each.
\textit{Top:} Number of linear regions.
\textit{Bottom:} Mean squared error loss.}
\label{fig:skipvsnoskip_linearregions}
\end{figure}

The results are shown in Fig.~\ref{fig:skipvsnoskip_linearregions}.
We observe that the networks with skip connections
consistently have more linear regions than the networks without 
skip connections throughout all stages of training. 
This result is surprising, as the number of neurons is the same
in both types of network, and so the upper bound on the number
of linear regions is the same. Ostensibly, one might expect there
to be no difference between the two networks in the early stages
of training, but one may expect a difference to appear after training.
Given the nature of this result, we perform additional hypothesis
testing in Subsect.~\ref{subsec:statisticaltest-randominit}
to verify this result for randomly-initialized networks.

\subsection{Comparison Upon Random Initialization}\label{subsec:statisticaltest-randominit}

In Subsect.~\ref{subsec:skipvsnoskip-training},
we found evidence that networks with skip connections may have more 
linear regions than equivalent networks without skip connections,
even without training. In this subsection, we perform hypothesis tests to verify this result.

For each test, we randomly initialize 50 neural networks without skip connections using Kaiming He initialization,
each with some given number of layers and neurons per layer
(as laid out in the \textit{``neural network parameters''} columns 
of Table~\ref{tab:linear-regions-tests}), and we initialize 50 networks with skip connections.
Taking $\mu_1$ to be the sample mean number of linear regions in the networks without skip connections and $\mu_2$ to be the sample mean with skip connections, we apply a one-tailed Mann-Whitney $U$ 
test with null hypothesis $\mu_1=\mu_2$ and alternative 
hypothesis $\mu_1 < \mu_2$, with significance level 0.05.
We use a $U$ test because the distribution of each sample is not Gaussian. 
Further, we find that the best-fitting distribution for each sample
is a shifted Gamma distribution.
Therefore, we also include goodness-of-fit tests for Gamma distributions
for each sample,
and we find no significant evidence to suggest that this Gamma distribution
is inappropriate.

\begin{table}[ht]
\centering
\scriptsize
\setlength{\tabcolsep}{1.5pt} 
\newcommand{\tabgroupmult}{8}
\renewcommand{\arraystretch}{1.8} 
\caption[Results of hypothesis tests for the mean number of linear regions 
         of neural networks with and without skip connections.]
{Results of hypothesis tests for the sample mean number of linear regions
of neural networks with and without skip connections.
For each row, 50 networks with skip connections
and 50 networks without skip connections are initialized.
An entry of e.g. $1\shortto3$ in the \textit{``Skip connections''}
column denotes that the output of the first layer is added to the
input of the third layer.
\textit{``KS test''} is a one-sample Kolmogorov-Smirnov test 
for the goodness-of-fit of a Gamma distribution.
One-tailed $U$-test compares the sample mean number of 
linear regions for networks with and without skip connections. 
}
\scalebox{0.95}{
\begin{tabular}{ccc @{\hskip\tabgroupmult\tabcolsep}
cc @{\hskip \tabgroupmult\tabcolsep} cc @{\hskip \tabgroupmult\tabcolsep}
cc @{\hskip 0pt} c @{\hskip 0pt}}
\toprule
\multicolumn{3}{@{}c@{\hskip \tabgroupmult\tabcolsep}}{\makecell{Neural network\\ parameters}}
& \multicolumn{2}{@{}c@{\hskip \tabgroupmult\tabcolsep}}{\makecell{KS test\\$p$-value}}
& \multicolumn{2}{@{}c@{\hskip \tabgroupmult\tabcolsep}}{\makecell{Sample mean no.\\of linear regions}} 
& \multicolumn{3}{@{}c@{\hskip \tabgroupmult\tabcolsep}}{\makecell{One-tailed\\Mann-Whitney $U$ test}} \\
\cmidrule(lr{\dimexpr \tabgroupmult\tabcolsep+0.5em}){1-3} 
\cmidrule(lr{\dimexpr \tabgroupmult\tabcolsep+0.5em}){4-5}
\cmidrule(lr{\dimexpr \tabgroupmult\tabcolsep+0.5em}){6-7}
\cmidrule(lr{\dimexpr \tabgroupmult\tabcolsep+0.5em}){8-10}
\makecell{No.\ of\\layers} & \makecell{Neurons\\per layer}
& \makecell{Skip\\connections} & \makecell{With\\skips}
& \makecell{Without\\skips} & \makecell{With\\skips}
& \makecell{Without\\skips} & \makecell{$U$-\\statistic}
& \makecell{$U$ test\\$p$-value} & \makecell{Reject null\\hypothesis} \\
\midrule
6 & 5 & \makecell{$1\shortto3$, $2\shortto4$,\\[-0.2em] $3\shortto5$} & 0.960 & 0.830 & 206.8 & 117.7 & 2237.5 & $1.0 \tabletimes 10^{^{-11}}$ & Yes \\
6 & 4 & \makecell{$1\shortto3$, $2\shortto4$,\\[-0.2em] $3\shortto5$} & 0.925 & 0.737 & 111.6 & 62.2 & 2161.5 & $3.4 \tabletimes 10^{^{-10}}$ & Yes \\
5 & 4 & $1\shortto3$, $2\shortto5$ & 0.801 & 0.709 & 96.7 & 53.3 & 2242.0 & $8.1 \tabletimes 10^{^{-12}}$ & Yes \\
4 & 4 & $1\shortto3$, $2\shortto5$ & 0.849 & 0.984 & 71.5 & 53.5 & 1804.5 & $1.3 \tabletimes 10^{^{-4}}$ & Yes \\
3 & 6 & $1\shortto3$ & 0.815 & 0.665 & 137.6 & 115.7 & 1711.5 & $1.5 \tabletimes 10^{^{-3}}$ & Yes \\
3 & 4 & $1\shortto3$ & 0.810 & 0.992 & 52.3 & 42.4 & 1728.0 & $9.9 \tabletimes 10^{^{-4}}$ & Yes \\
\bottomrule
\end{tabular}
}
\label{tab:linear-regions-tests}
\end{table}

The results are given in Table~\ref{tab:linear-regions-tests}.
In each test, we obtain small $p$-values 
and reject the null hypothesis.
These tests provide us with strong evidence that the average number
of linear regions for untrained networks with skip connections
is greater than that of equivalent untrained networks
without skip connections.


\subsection{Analysis of Results}\label{subsec:skipvsnoskip-analysis}

In Subsect.~\ref{subsec:skipvsnoskip-training},
we observed that neural networks with skip connections have more
linear regions throughout training than regular networks,
and achieve a greater maximum number of skip connections
when training finishes. 
This indicates that networks with skip connections produce 
a more advanced output map with more facets, 
and are therefore more expressive than regular networks. 

One explanation of this observation is that layers directly 
following skip connections have more flexibility to partition 
either the space from the previous layer, or partition the space
from the ``skipped'' layers. 
More precisely, suppose we have a skip connection from layer $k$
to layer $\ell+1$, so that the input to the $(\ell+1)$-th layer 
is $\nu^{(k)}(x)+\nu^{(\ell)}(x)$. 
Then layer $(\ell+1)$ introduces hyperplanes with equations 
$G^{(\ell+1)}(x)=H^{(\ell+1)}(x)$, and by (\ref{eq:recurrenceskip2}), this is equivalent to:
 
\begin{equation}\label{eq:skip_hyperplane}
   G^{(\ell)}(x)+A_{-}\nu^{(k)}(x) = H^{(\ell)}(x)+A_{+}\nu^{(k)}(x)
\end{equation} 
On the other hand, a neural network without this skip connection 
would introduce hyperplanes at $G^{(\ell)}(x) = H^{(\ell)}(x)$.
Therefore, when training a neural network, each skip connection provides the optimization algorithm
with an opportunity to introduce a hyperplane as a function of the 
outputs of both layers $\ell$ and $k$, rather than layer $\ell$ alone,
leading to more linear regions.

In Subsect.~\ref{subsec:statisticaltest-randominit}, 
we found evidence that the average 
number of linear regions for untrained neural networks 
with skip connections is greater than that of equivalent networks
without skip connections. 
This shows that adding skip connections makes a network 
inherently more expressive, rather than simply improving its ability
to be optimized with gradient descent methods. 
To reason this out, consider that a layer that uses a skip connection
as its input introduces a hyperplane at (\ref{eq:skip_hyperplane})
--- the matrices $A_{+}$ and $A_{-}$ have random entries,
and so a random multiple of $\nu^{(k)}$ is added to the hyperplane equation.
This aligns the direction of the new hyperplane more with $\nu^{(k)}$,
making it more likely to intersect each existing linear region and to split them into two. 
This result may be possible to formally prove by using random variables
to represent the coefficients of the boundaries of linear regions,
which we foresee as a future research direction.

\section{Application: Caching Values in Neural Networks}\label{sec:caching}

In this section, we propose a method to compress nonlinear neural networks
into as a single layer (which we originally envisioned in \cite{JV24}), given by Algorithm~\ref{algorithm:caching}.

\begin{algorithm}[htb]\caption{Algorithm to make predictions 
and cache observed linear regions.}\label{algorithm:caching}
\begin{algorithmic}[1]
\Function{Predict}{$x$}
    \State Find linear the region $v \in \{-1,1\}^\texttt{num\_neurons}$ in which $x$ lies.

    \Comment{This can be achieved by checking the inequalities that define each linear region.} 
    \If{$v$ is not in the cache}
        \State Use recurrence relations defined in (\ref{eq:decomposedrecurrence})
	to find the matrix $F^{(L)}_A$ and vector $f^{(L)}_b$ 
	s.t.\ the neural network map is $F^{(L)}_A x + f^{(L)}_b$ on the given linear region.
        \State Append the key-value pair $\big(v$ : $(F^{(L)}_A, f^{(L)}_b)\big)$ to the cache
    \EndIf
    \State Use the cache to look up $F^{(L)}_A, f^{(L)}_b$ for the current linear region
    \State \Return $F^{(L)}_A x + f^{(L)}_b$
\EndFunction
\end{algorithmic}
\end{algorithm}

As a caching algorithm, the speedup we observe depends entirely 
on how often each new instance arrives pre-cached,
which is highly dependent on both the complexity of the neural network,
as well as the data set on which the network has been trained.
This raises the question of how often trained models cache instances,
and whether networks with skip connections
cache data at a different rate than networks 
without skip connections. In the remainder of this section, we seek answers to 
these questions by performing experiments with our caching algorithm.
In Subsect.~\ref{subsec:caching-experiment1},
we train classifier models of various sizes on three different data sets.
In Subsect.~\ref{subsec:caching-experiment2},
we examine the effect of using skip connections
on the rate at which data instances are cached.

\subsection{Caching Experiment on Various Data Sets}\label{subsec:caching-experiment1}

To examine how often cached results are used,
we train models of varying sizes on the following three different data sets: 
\begin{enumerate*}[label=\itshape\alph*\upshape)]
\item MNIST data set of handwritten digits,
containing 60,000 training instances and 10,000 test instances, 
\item CIFAR-10 data set \cite{Kri09}, containing 50,000 training instances
and 10,000 testing instances (for computational tractability,
we desaturate the 3-channel RGB images into 1-channel grayscale images),
\item Street View House Number (SVHN) data set~\cite{NWCBWN11},
containing 73,257 training instances and 26,032 testing instances.
\end{enumerate*}





For each of the three data sets, we train a network with 4 layers
and 8 neurons per layer, a network with 3 layers and 16 neurons per layer,
and a network with 3 layers and 32 neurons per layer.
Each network is trained on each data set with the Adam optimizer
for 5 epochs with learning rate $\eta=5\times10^{-4}$.

After training, we cache all linear regions that contain 
at least one training instance by storing a string with a 1 or 0 for each neuron, 
indicating whether the neuron's pre-activation function is positive
or negative for the given training instance.
For each testing instance, we then find the linear region in which
the instance lies, and check whether any training instance lies in the same region.

\begin{figure}[htb]
\centering
\setlength\tabcolsep{2pt}
\renewcommand{\arraystretch}{0} 
\begin{tabular}{ccc}
\includegraphics[width=0.28\textwidth]{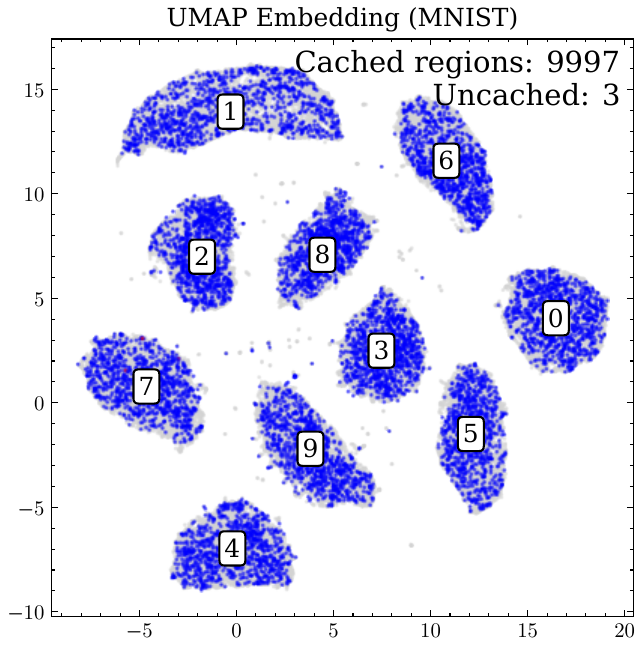}
& \includegraphics[width=0.28\textwidth]{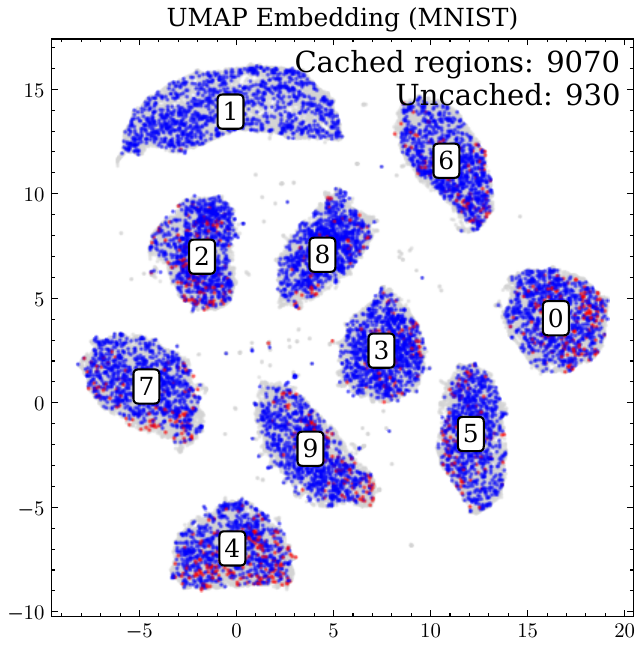}
& \includegraphics[width=0.28\textwidth]{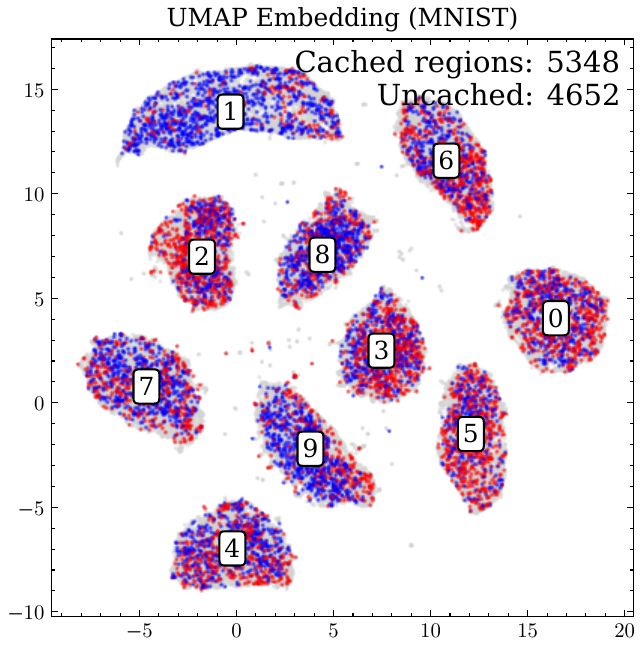} \\
\includegraphics[width=0.28\textwidth]{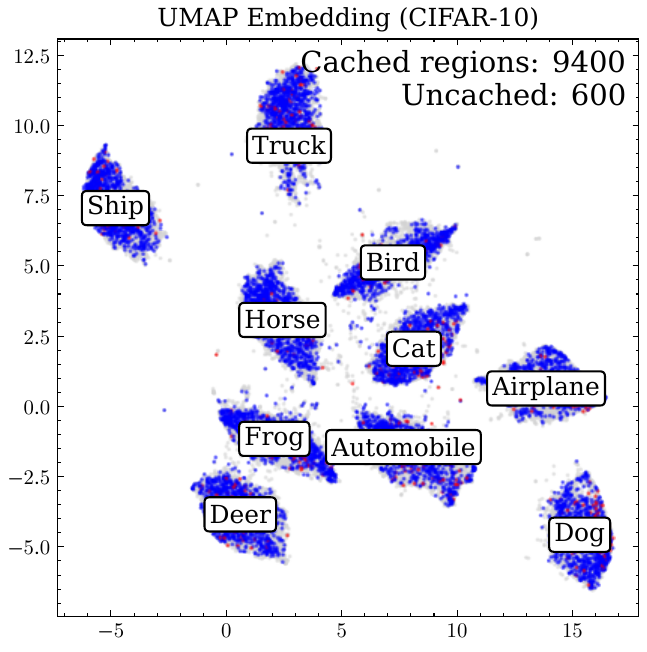}
& \includegraphics[width=0.28\textwidth]{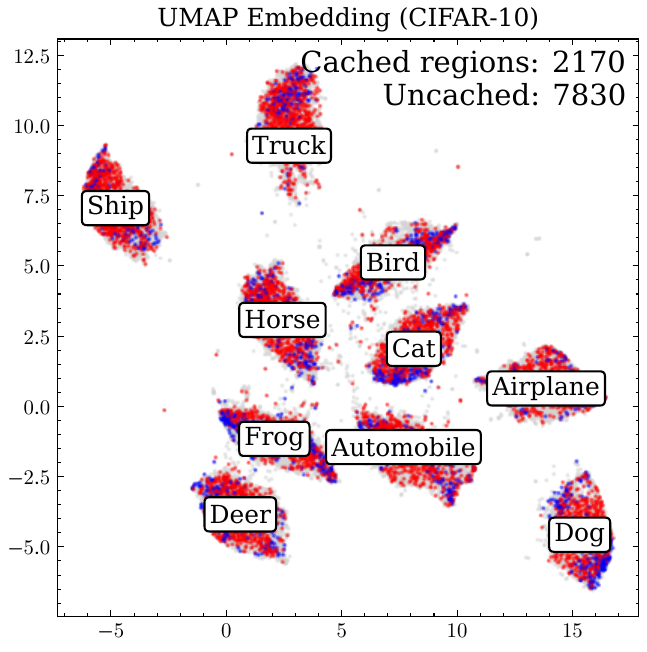}
& \includegraphics[width=0.28\textwidth]{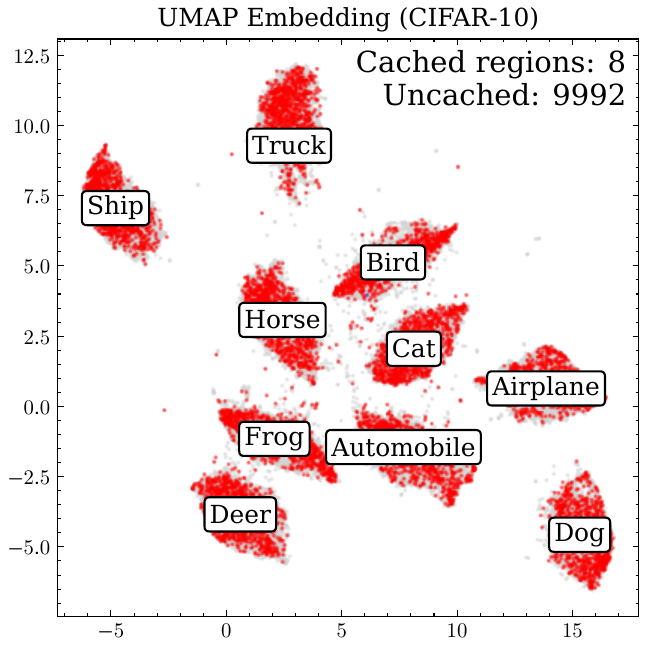} \\
\includegraphics[width=0.28\textwidth]{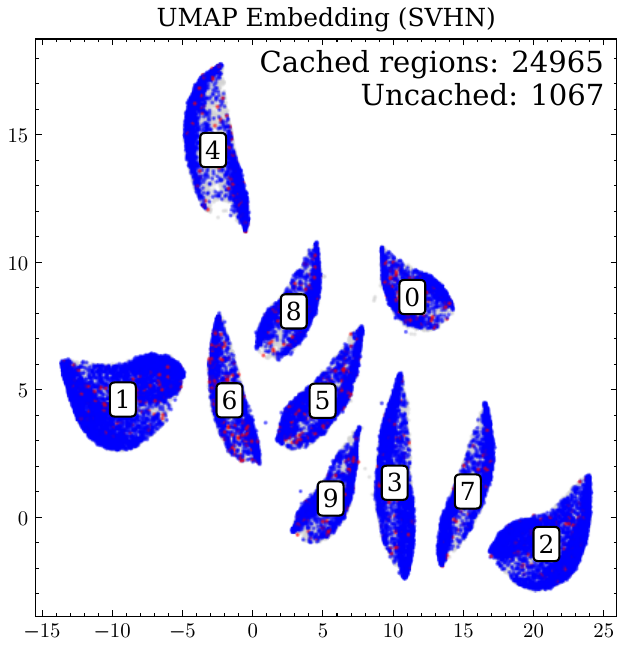}
& \includegraphics[width=0.28\textwidth]{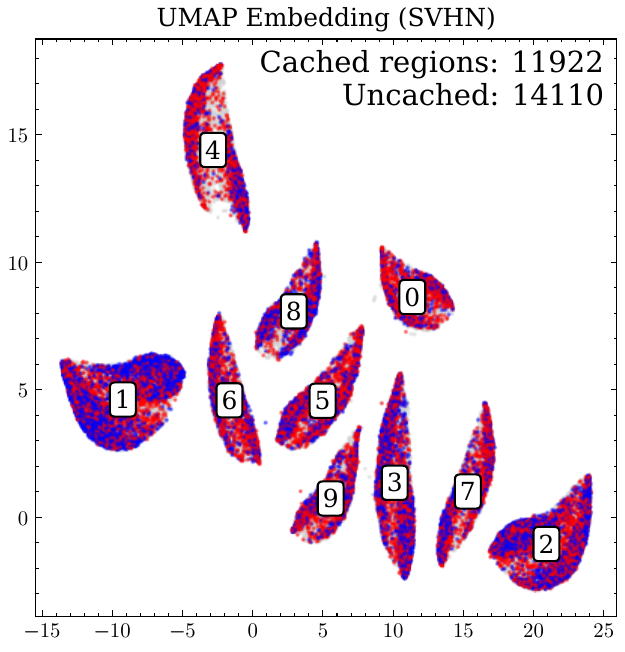}
& \includegraphics[width=0.28\textwidth]{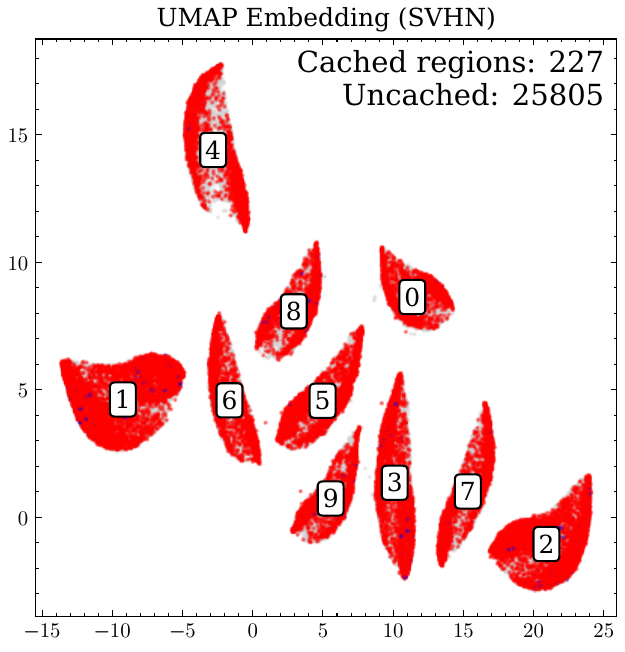} 
\end{tabular}
\caption[UMAP embeddings showing rates of caching
         for various data sets and model sizes.]
{UMAP embeddings showing rates of caching in the test set for each of;
MNIST \textit{(top row)}, CIFAR-10 \textit{(middle row)}, and SVHN \textit{(bottom row)}. 
Each column represents a different model size; 4 layers and 8 neurons 
per layer \textit{(left column)}, 3 layers and 16 neurons 
per layer \textit{(middle column)}, and 3 layers and 32 neurons 
per layer \textit{(right column)}.
Blue points denote test data in the same linear region as at least 1 training set instance, while red points are \emph{not} in the same region as any training instance. Gray points represent training data. 
Labels show the mean coordinates of each class.}
\label{fig:pca-umap}
\end{figure}


The results are shown in Fig.~\ref{fig:pca-umap}.
For easier visualization,
we use Uniform Manifold Approximation and Projection (UMAP) \cite{MHSG18}
to plot the data in 2D.
Instances in the test set that lie
in the same linear region as at least one training instance
are colored in blue, while the rest of the test set is in red.
For simplicity, let us refer to these types of instances 
as cached and uncached, respectively. 

Based on Fig.~\ref{fig:pca-umap}, we can make several observations.
\begin{enumerate*}[label=\itshape\alph*\upshape)]
\item Rates of caching are not uniform across each class. 
For instance, in the plot corresponding to the 3-layer, 
32-neuron-per-layer model on MNIST (top-right position
of Fig.~\ref{fig:pca-umap}), we observe a mix of cached and uncached 
test instances, and we see that the labels ``9'' and ``1'' have
high rates of caching, while ``2'' and ``3'' are much lower.
\item From the middle column of Fig.~\ref{fig:pca-umap},
we see that uncached instances tend to appear further away 
from class centroids, while the cached instances are usually 
closer to the centroids. This shows that there may be potential
for this method to be able to detect out-of-distribution data. 
Such a method could involve mapping
out linear regions as a graph, and counting
the path length to the nearest in-distribution element.
\item The proportion of cached elements depends on the number
of neurons in the network. For larger networks (right column
of Fig.~\ref{fig:pca-umap}), we have more linear regions,
so it is less likely that any two data points lie in the same region.
To use this technique on large networks, knowledge distillation 
techniques may be needed.
\item The data set choice affects the rate of caching. 
This is most evident when comparing CIFAR-10 and MNIST,
which have the same number of testing instances,
but more test data is cached
on MNIST than for CIFAR-10, which may be due to the higher
relative difficulty of CIFAR-10.
\end{enumerate*}

\subsection{Caching Experiment With and Without Skip Connections}\label{subsec:caching-experiment2}




Using a similar setup as in Sect.~\ref{subsec:caching-experiment1},
we can also compare networks with and without skip connections. 
For each model-dataset pair as given in Fig.~\ref{fig:pca-umap}, we conduct a trial where we train 50 networks with skip connections, and 50 networks without skip connections, and record the number of cached values. 
For each trial, take $\mu_1$ to be the sample mean number of cached
instances of the networks without skip connection and $\mu_2$ to be
the sample mean of the networks with skip connections
--- we apply a one-tailed Mann-Whitney $U$ test with null hypothesis
$\mu_1=\mu_2$ and alternative hypothesis $\mu_1 > \mu_2$,
with significance level 0.05 for each test. The results of these hypothesis tests are given
in Table~\ref{tab:cached-tests}.

\begin{table}[ht]
\scriptsize
\setlength{\tabcolsep}{1.8pt} 
\newcommand{\tabgroupmult}{2}
\renewcommand{\arraystretch}{2.5} 
\caption[Results of hypothesis tests for the sample mean number
         of cached values found in the testing set for various data sets.]
{Results of hypothesis tests for the sample mean number of cached values
found in the testing set for various data sets.
For each row, 50 networks with skip connections
(as in \textit{``Skip connections''} column, 
which uses the same format as Table~\ref{tab:linear-regions-tests})
are trained,
and 50 networks without skip connections are trained.
All networks are trained for 5 epochs on MNIST, CIFAR-10,
or SVHN, respectively.
Using Algorithm~\ref{algorithm:caching}, the number of cached 
test instances is recorded for each trained model.
A one-tailed Mann-Whitney $U$ test is applied to compare the means
with and without skip connections.}
\scalebox{0.9}{
\begin{tabular}{r @{\hskip\tabgroupmult\tabcolsep} 
ccc @{\hskip\tabgroupmult\tabcolsep}
cc @{\hskip \tabgroupmult\tabcolsep}
cc @{\hskip \tabgroupmult\tabcolsep}
cc @{\hskip 0pt} c @{\hskip 0pt}}
\toprule
\multirow{2}{*}{\makecell{Data\\set}}
& \multicolumn{3}{@{}c@{\hskip \tabgroupmult\tabcolsep}}{\makecell{Neural network\\parameters}}
& \multicolumn{2}{@{}c@{\hskip \tabgroupmult\tabcolsep}}{\makecell{Mean accuracy\\on test set}}
& \multicolumn{2}{@{}c@{\hskip \tabgroupmult\tabcolsep}}{\makecell{Mean no.\ of cached\\values in test set}}
& \multicolumn{3}{@{}c@{\hskip \tabgroupmult\tabcolsep}}{\makecell{One-tailed\\Mann-Whitney $U$ test}} \\
\cmidrule(lr{\dimexpr \tabgroupmult\tabcolsep+0.5em}){2-4}
\cmidrule(lr{\dimexpr \tabgroupmult\tabcolsep+0.5em}){5-6}
\cmidrule(lr{\dimexpr \tabgroupmult\tabcolsep+0.5em}){7-8}
\cmidrule(lr{\dimexpr \tabgroupmult\tabcolsep+0.5em}){9-11}
& \makecell{No.\ of\\layers} & \makecell{Neurons\\per layer}
& \makecell{Skip\\connections} & \makecell{With\\skips}
& \makecell{Without\\skips} & \makecell{With\\skips}
& \makecell{Without\\skips} & \makecell{$U$-\\statistic}
& \makecell{$U$ test\\$p$-value} & \makecell{Reject null\\hypothesis} \\
\midrule
\multirow{3}{*}[0.75em]{MNIST} & 4 & 8 & $1\shortto3$, $2\shortto4$ & 83.8\% & 78.7\% & 9966.4 & 9970.0 & 1245.5 & $0.489$ & No\\[-1.2em]
 & 3 & 16 & $1\shortto3$ & 91.8\% & 91.3\% & 9211.6 & 9349.2 & 997.5 & 0.041 & Yes\\[-1.2em]
 & 3 & 32 & $1\shortto3$ & 94.4\% & 93.9\% & 2140.3 & 3200.6 & 397.5 & $2.1 \tabletimes 10^{^{-9}}$ & Yes\\[-0.6em]
\multirow{3}{*}[0.75em]{CIFAR-10} & 4 & 8 & $1\shortto3$, $2\shortto4$ & 30.3\% & 29.2\% & 8965.6 & 9125.4 & 850.5 & $3.0 \tabletimes 10^{^{-3}}$ & Yes\\[-1.2em]
 & 3 & 16 & $1\shortto3$ & 35.8\% & 35.2\% & 1702.2 & 2436.5 & 314.0 & $5.6 \tabletimes 10^{^{-11}}$ & Yes\\[-1.2em]
 & 3 & 32 & $1\shortto3$ & 39.0\% & 39.0\% & 4.1 & 7.3 & 485.0 & $5.7 \tabletimes 10^{^{-8}}$ & Yes\\[-0.6em]
\multirow{3}{*}[0.75em]{SVHN} & 4 & 8 & $1\shortto3$, $2\shortto4$ & 52.5\% & 48.8\% & 24806.9 & 24982.8 & 877.0 & $5.1 \tabletimes 10^{^{-3}}$ & Yes\\[-1.2em]
 & 3 & 16 & $1\shortto3$ & 64.9\% & 63.4\% & 10273.2 & 12532.9 & 304.5 & $3.6 \tabletimes 10^{^{-11}}$ & Yes\\[-1.2em]
 & 3 & 32 & $1\shortto3$ & 72.6\% & 71.6\% & 87.3 & 174.7 & 289.5 & $1.8 \tabletimes 10^{^{-11}}$ & Yes\\[-0.4em]
\bottomrule
\end{tabular}
}
\label{tab:cached-tests}
\end{table}

From Table~\ref{tab:cached-tests}, we reject the null hypothesis
for all but one model-dataset pair\footnote{We did not reject 
the null hypothesis when using the 4-layer network on MNIST. 
Here, almost all test instances were cached regardless of the presence
of skip connections ($\frac{9966.4}{10000}\approx 99.7\%$ vs $\frac{9970}{10000}\approx 99.7\%$), giving nearly equal population means.}.
We also observe that the networks with skip connections have greater or equal
average accuracy than the networks without skip connections, 
which aligns with conventional knowledge.
When considered together with the fact that this accuracy comes
from uncached data, this indicates that introducing skip
connections to a neural network allows it to generalize to unseen
data more effectively.
Combined with our observation in Sect.~\ref{sec:linearregions-skipvsnoskip}
that neural networks with skip connections have more linear regions,
this shows that networks with skip connections do not simply 
introduce linear regions around individual training instances,
but instead produce regions that are representative of underlying patterns.

\section{Conclusion}\label{sec:tropical-conclusion}

Recurrence relations form the basis for algorithms to
compute all linear regions of a neural network,
generalized to work for networks with skip connections.
Visualizations of those linear regions help to understand
how patterns in the training data may lead to overfitting.
Caching the linear transformations allows for faster predictions.
Through experimentation, we found that
skip connections allow a model to both be more expressive
by creating more advanced output maps, 
while also ensuring that these maps can generalize to new data.




\bibliographystyle{plain}

\end{document}